%% file: metafed-tnnls.tex
\documentclass[lettersize,journal]{IEEEtran}
\usepackage{cite}
\usepackage{amsmath,amssymb,amsfonts,amsthm}
\usepackage{algorithm}
\usepackage{algorithmicx}
\usepackage[noend]{algpseudocode}
\usepackage{graphicx}
\usepackage{textcomp}
\usepackage{xcolor}
\usepackage{url}
\usepackage{hyperref}
\usepackage{multicol}
\usepackage{multirow}
\usepackage{booktabs}
\usepackage{subfigure}
\usepackage{todonotes}
\usepackage{xcolor}
\usepackage{xspace}
\usepackage[normalem]{ulem}

\theoremstyle{plain}

\theoremstyle{definition}

\theoremstyle{remark}

\newcommand{\algorithmname}{Algorithm}
\newcommand{\equationname}{Eq.}

\newcommand{\sectionname}{Sec.}
\newcommand{\paragraphname}{Para.}

\newcommand{\lw}[1]{{\color{black}{#1}}}

\newcommand{\method}{{MetaFed}\xspace}

\begin{document}

\title{MetaFed: Federated Learning among Federations with Cyclic Knowledge Distillation for Personalized Healthcare}

\author{Yiqiang~Chen,
        Wang~Lu,
        Xin Qin, 
        Jindong~Wang,
        and~Xing~Xie,~\IEEEmembership{Fellow,~IEEE}
        \thanks{Yiqiang Chen, Wang Lu and Xin Qin are with Institute of Computing Technology, Chinese Academy of Sciences. E-mail: \{luwang, yqchen\}@ict.ac.cn.  Correspondence to Yiqiang Chen.}
        \thanks{Jindong Wang and Xing Xie are with Microsoft Research Asia. E-mail: \{jindong.wang, xingx\}@microsoft.com.}
        }


        
\markboth{Preprint}%
{Shell \MakeLowercase{\textit{et al.}}: A Sample Article Using IEEEtran.cls for IEEE Journals}


\maketitle

\begin{abstract}
Federated learning has attracted increasing attention to building models without accessing the raw user data, especially in healthcare.
In real applications, different federations can seldom work together due to possible reasons such as data heterogeneity and distrust/inexistence of the central server.
In this paper, we propose a novel framework called \textbf{MetaFed} to facilitate trustworthy FL between different federations.
\method obtains a personalized model for each federation without a central server via the proposed Cyclic Knowledge Distillation.
Specifically, \method treats each federation as a meta distribution and aggregates knowledge of each federation in a \emph{cyclic} manner. 
The training is split into two parts: common knowledge accumulation and personalization. 
Comprehensive experiments on seven benchmarks demonstrate that \method without a server achieves better accuracy compared to state-of-the-art methods (e.g., 10\%+ accuracy improvement compared to the baseline for PAMAP2) with fewer communication costs. 
More importantly, \method shows remarkable performance in real healthcare-related applications.

\end{abstract}


\section{Introduction}

\IEEEPARstart{M}achine learning, especially deep learning, has become prominent in people's daily lives~\cite{lu2022local,lu2022semantic,he2021locality}.
It is applied to many health-related fields such as human activity recognition~\cite{lu2021cross}, medical images~\cite{li2021medical}, and other fields~\cite{DBLP:conf/kdd/MaYLXS21, aguiar2022learning, liu2022contribution}.
However, with the increasing awareness of data privacy and security, some countries and organizations released policies to prevent data leakage~\cite{inkster2018china,voigt2017eu}.
In this situation, federated learning (FL)~\cite{yang2019federated} was proposed and becomes increasingly popular.



Google~\cite{mcmahan2017communication} proposed the first FL algorithm called FedAvg to aggregate clients' information.
FedAvg replaces direct data exchanges with model parameter communication to preserve data privacy. 
\cite{yang2019federated} offers a review of federated learning and it groups federated learning into three categorizes, including horizontal federated learning, vertical federated learning, and transfer federated learning.
Although FedAvg achieves promising performance in many applications, it may not be feasible in more challenging situations.
For example, when meeting non-iid data, FedAvg will suffer from slow convergence and low accuracy.
Therefore, some federated learning methods, e.g. FedBN~\cite{li2021fedbn}, are proposed to solve these concrete problems.

\begin{figure}[t!]
	\centering
	\includegraphics[width=0.5\textwidth]{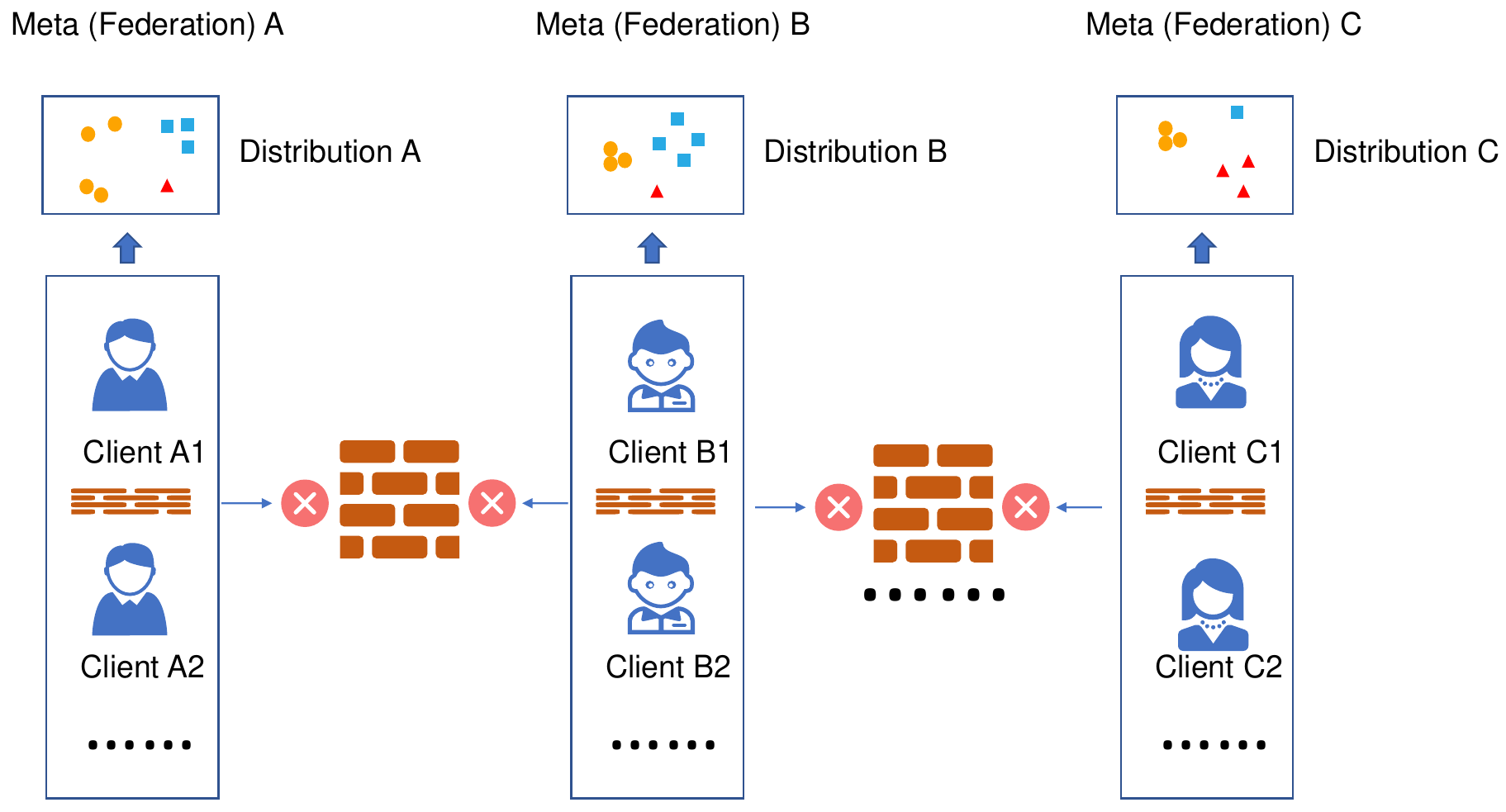}
	\caption{Issues in meta federated learning.}
	\label{fig:mfliss}
\end{figure}

In reality, the situations can be more difficult. 
Medical institutions may be grouped into multiple federations and no higher-level governing organizations exist.
Consider a real example.
Several hospitals form a federation which is called the inner-hospital federation while several medical companies, providing service to consumers at home, form another federation which is called the outer-hospital federation.
Obviously, we can perform FedAvg inside both two federations.
But, how can we combine these two federations?
There is no higher-level server and existing methods, e.g. FedAvg and FedBN, all fail in this situation.
\figurename~\ref{fig:mfliss} gives an abstract summary of this real challenging situation.

As shown in \figurename~\ref{fig:mfliss}, a certain number of clients form a federation, and different federations are independent enough that do not use a central server, but communicate with each other instead.
Inside each federation, different FL algorithms can be used to train a model.
However, it remains unclear how to build personalized FL models outside the federations, i.e., FL among different federations.
Moreover, data heterogeneity widely exists in these federations.
We view each federation as a meta distribution and view the problem in this situation as meta federated learning.\footnote{We use meta and federation interchangeably.}

In this article, we propose \textbf{\method}, a meta federated learning framework for cross-federation federated learning.
We focus on inter-federation federated learning in this paper and each federation can be viewed as an independent individual.
To implement \method, we propose a cyclic knowledge distillation method.
\method can solve data islanding and data statistical heterogeneity without requiring a server or sacrificing user privacy.
Specifically, \method consists of two stages, common knowledge accumulation stage and personalization stage. 
In the first stage, the model trained on the previous meta federation serves as the teacher for the next federation and knowledge distillation (KD)~\cite{hinton2015distilling,romero2014fitnets} aims to make use of the common information. 
Several rounds with fixed hyperparameters for knowledge distillation are performed to ensure enough common knowledge. 
In personalization stage, we utilize KD with adapted hyperparameters to obtain personalized model for each federation. 
Through knowledge distillation, it can not only acquire common knowledge among federations but also cope with feature shifts and label shifts. 
Moreover, \method is extensible and can be deployed to many applications.
The code for \method is released at \url{https://github.com/microsoft/PersonalizedFL}.

Our contributions are as follows.
\begin{enumerate}
    \item We propose \method, a novel meta federated learning framework via cyclic knowledge distillation for healthcare, which can accumulate common information from different federations without compromising privacy security, and achieve personalized models for each federation through adapted knowledge distillation.
    \item Comprehensive experiments on image and time-series datasets illustrate that \method has remarkable performance in each federation without a server compared to state-of-the-art methods. Moreover, \method reduces the number of rounds, thus saving communication costs.
    \item Two real health-related applications, COVID-19 under label shifts and PD Tremor under feature shifts, demonstrate that \method can also achieve noteworthy performance in reality and \method can be directly utilized in real personalized federated learning.
    \item \method is extensible and can be applied in many healthcare applications, which means it can work well in many circumstances. We can even replace the knowledge distillation with some other incremental learning methods for specific applications. 
\end{enumerate}

The remainder of this paper is organized as follows.
We introduce related work in \sectionname~\ref{sec-relate}.
In \sectionname~\ref{sec-method}, we elaborate the proposed method and offer a clear summary.
Then, the experimental implementations and results are presented to demonstrate the superiority of \method in \sectionname~\ref{sec:exp}.
Moreover, two real applications are provided in \sectionname~\ref{sec-realword} to illustrate that our method can work well in reality.
Finally, we conclude the paper and provide some future directions in \sectionname~\ref{sec-conclus}. 

\section{Related Work}
\label{sec-relate}
\subsection{Machine Learning for Healthcare}
Machine learning has shown its superior in many application fields, especially in healthcare~\cite{paluru2021anam,ding2021deepkeygen,yu2022healthnet}.
From the simplest application where human daily activities are recognized~\cite{wang2019deep} to the most difficult one where brain tumors can be classified~\cite{9129779}, machine learning can exert its effectiveness.
In a common framework, it often first collects lots of labeled data, then extracts features, and finally trains a classifier. 
Unfortunately, the quality and number of collected labeled data in the first step often have a significant impact on the application. 
In many real applications, especially healthcare-related ones, obtaining centralized data is impossible. 
Although multiple persons or organizations generate a large amount of data, few of them are willing to share their private data~\cite{tan2022towards}. 
Moreover, a large number of regulations have emerged, such as~\cite{inkster2018china,voigt2017eu}, to protect data privacy and security. 
These factors make data form separate data islands where learning a model with aggregated data cannot work. 
\lw{In addition, heterogeneous data statistics can be another issue that seriously hinders the development of machine learning in healthcare~\cite{vogenberg2010personalized}.}
Therefore, how to learn a robust model without directly exchanging data becomes a trend.

\subsection{Federated Learning}
To make full use of data in different separate clients and protect data privacy and security simultaneously, Google~\cite{mcmahan2017communication} first proposed FedAvg to train machine learning models via aggregating distributed mobile phones' data with exchanging model parameters instead of directly exchanging data.
FedAvg can work well with data islanding problems in many applications although it is simple.
Subsequently, Yang et al.~\cite{yang2019federated} wrote the first survey of FL research.

Federated learning has attracted growing attention in many applications. 
And the traditional and simple FedAvg cannot satisfy complicated realistic scenes. 
When meeting data statistical heterogeneity, FedAvg may converge slowly and acquire large amounts of communication cost. 
Moreover, since only a shared global model is obtained, the model may degrade when predicting in personalized clients.
Some work tries to cope with these problems.
FedProx~\cite{li2018federated} added a proximal term to FedAvg which referred to the global model and allowed slight differences when training local models. 
Yu et al.~\cite{yu2020salvaging} combined three traditional adaptation techniques: fine-tuning, multi-task learning, and knowledge distillation into federated models.
Most recently, FedBN~\cite{li2021fedbn} tried to cope with feature shifts among clients by preserving local batch normalization parameters which can represent data distributions to some extent.
\lw{FED-ROD~\cite{chenbridging} paid attention to both generalization and personalization.} 
Some other work made an effort to utilize personalization federated learning in the healthcare field~\cite{chen2020fedhealth,lu2022personalized}. 
\cite{chen2020fedhealth} proposed a federated transfer learning framework that needs some sharing data while \cite{lu2022personalized} proposed FedAP which could achieve personalized performance via aggregating with clients' similarities.
However, these methods need a server and have some limits in communication costs.

In this situation where no server exists, FedAvg even cannot be implemented.
Sequential training may be a reasonable solution.
\cite{kopparapu2020fedfmc} proposed FedFMC that dynamically forked devices into updating different global models and merged models in a lifelong way.
\cite{zaccone2022speeding} leveraged the sequential training of subgroups of heterogeneous clients to emulate the centralized paradigm.
\cite{zeng2022heterogeneous} assigned clients to homogeneous groups to minimize the overall distribution divergence among groups.
These methods still rely heavily on parallel federated learning where sequential training round style with only one round and no closed loop is just an aid.  

\begin{figure}[t!]
	\centering
	\includegraphics[width=0.5\textwidth]{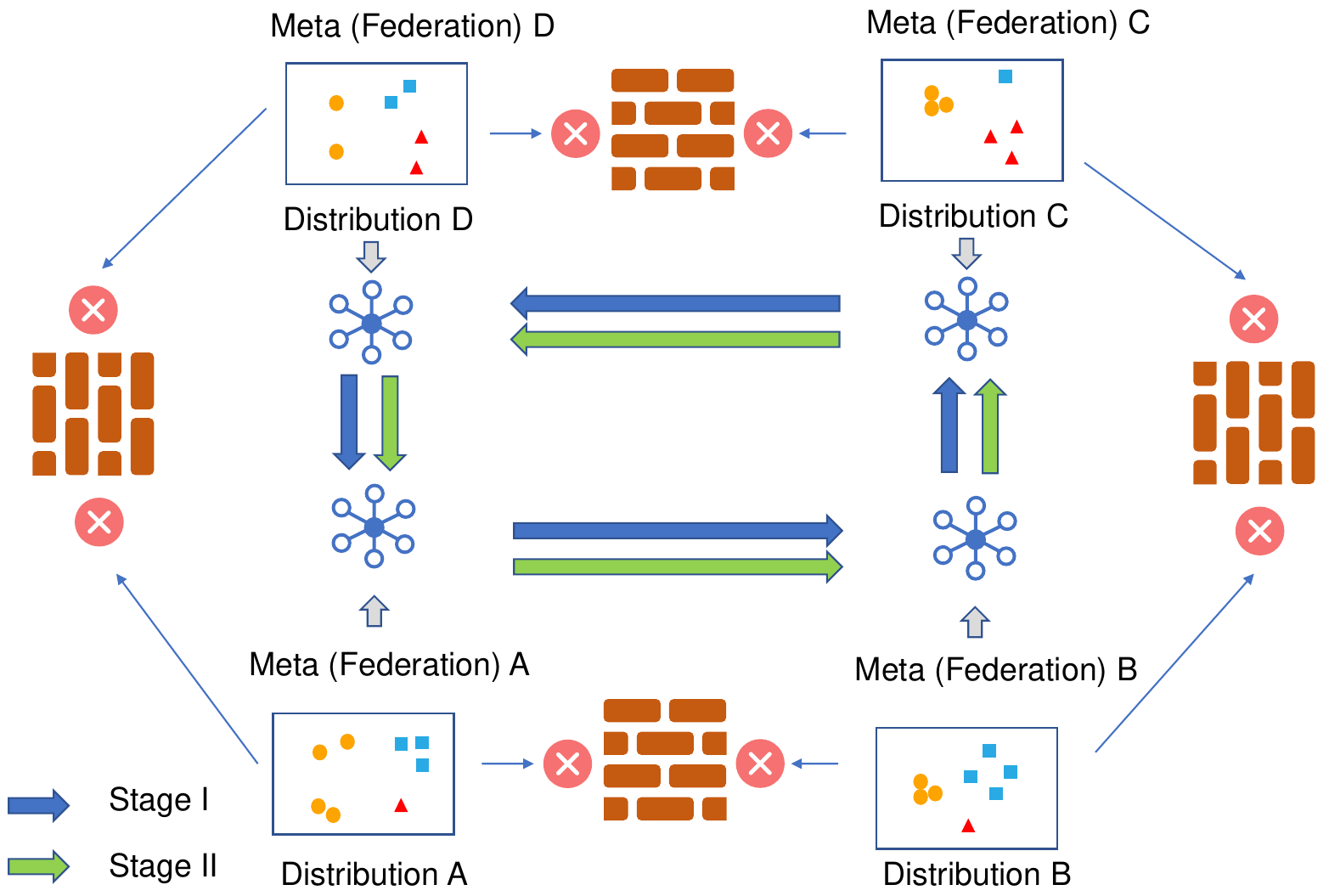}
	\caption{The structure of \method for federated learning among federations. Stage I is the common knowledge accumulation stage where the model is sent after local training. Stage II is the personalization stage where the common knowledge model is sent before local training.}
	\label{fig:frame1}
\end{figure}

Some other work, e.g. \cite{roy2019braintorrent,rieke2020future,li2021fedh2l,warnat2021swarm}, communicated in a peer-to-peer environment without a server. 
BrainTorrent~\cite{roy2019braintorrent} presented a highly dynamic peer-to-peer environment, where all centers directly interacted with each other without depending on a central body.
It seemed disorderly and chaotic, and it required lots of communication costs.
Nicola Rieke et al.~\cite{rieke2020future} considered key factors to federated learning while FedH2L~\cite{li2021fedh2l} utilized mutual distillation to exchange posteriors on a shared seed set between participants in a decentralized manner.
\lw{Swarm learning~\cite{warnat2021swarm} applied this style into the real decentralized and confidential clinical machine learning,} 
However, these methods often require large communication costs, and few are designed for personalization federation federated learning.
No work pays attention to proposing a new paradigm for personalized federated learning among federations without a server.

\subsection{Knowledge Distillation}
Knowledge distillation has been a well-known technique to transfer knowledge since birth~\cite{hinton2015distilling}. 
In the original version, the knowledge was transferred by mimicking the outputs of the teacher model on the same data. 
Later, besides imitating outputs, some work demonstrated that feature imitation could also guide the student model training~\cite{romero2014fitnets}.
\lw{
A more detailed description of knowledge distillation can be found in \cite{zhang2021survey,meng2021knowledge}.}
Nowadays, as a common technique, knowledge distillation is also applied to federated learning~\cite{usmanova2021distillation,afonin2021towards}.
Though mimicking the global model and the local previous model, different implementations can be flexibly applied to different situations.

\section{Method}
\label{sec-method}
\subsection{Problem Formulation}
In a personalized federated learning among federations setting, $N$ different federations, denoted as $\{ F_1, F_2, \cdots, F_N \}$, have data, denoted as $\{ \mathcal{D}_1, \mathcal{D}_2, \cdots, \mathcal{D}_N \}$ with different distributions, which means $P(\mathcal{D}_i) \neq P(\mathcal{D}_j)$.
For simplicity, we only study the case where the input and output spaces are the same, i.e. $\mathcal{X}_i = \mathcal{X}_j, \mathcal{Y}_i = \mathcal{Y}_j, \forall i\neq j$.
Each dataset, $\mathcal{D}_i = \{ (\mathbf{x}_{i,j}, y_{i,j}) \}_{j=1}^{n_i}$, consists of three parts, a train dataset $\mathcal{D}_i^{train} = \{ (\mathbf{x}_{i,j}^{train}, y_{i,j}^{train}) \}_{j=1}^{n_i^{train}}$, a validation dataset $\mathcal{D}_i^{valid} = \{ (\mathbf{x}_{i,j}^{valid}, y_{i,j}^{valid}) \}_{j=1}^{n_i^{valid}}$ and a test dataset $\mathcal{D}_i^{test} = \{ (\mathbf{x}_{i,j}^{test}, y_{i,j}^{test}) \}_{j=1}^{n_i^{test}}$. 
We have $n_i = n_i^{train} + n_i^{valid} + n_i^{test}$ and $\mathcal{D}_i = \mathcal{D}_i^{train} \cup \mathcal{D}_i^{valid} \cup \mathcal{D}_i^{test}$.
We aim to combine information of all federations without data exchange to learn a good model $f_i$ for each federation on its local dataset $\mathcal{D}_i$:
\begin{equation}
    \min_{\{f_k\}_{k=1}^N} \frac{1}{N} \sum_{i=1}^N \frac{1}{n_{i}^{test}} \sum_{j=1}^{n_i^{test}} \ell(f_i(\mathbf{x}_{i,j}^{test}), y_{i,j}^{test}),
    \label{eqa:goal}
\end{equation}
where $\ell$ is a loss function.

\subsection{Overview of \method}
Consider the union of different federations where there is no server among them and distribution shifts exist. 
How to make them communicate equally without any other governors and share common knowledge without direct data exchange is the key. 
\method aims to accumulate common knowledge and preserve personalized information without compromising data privacy and security via knowledge distillation in a cyclic way.
\figurename~\ref{fig:frame1} gives an overview.

\begin{figure}[t!]
	\centering
	\includegraphics[width=0.5\textwidth]{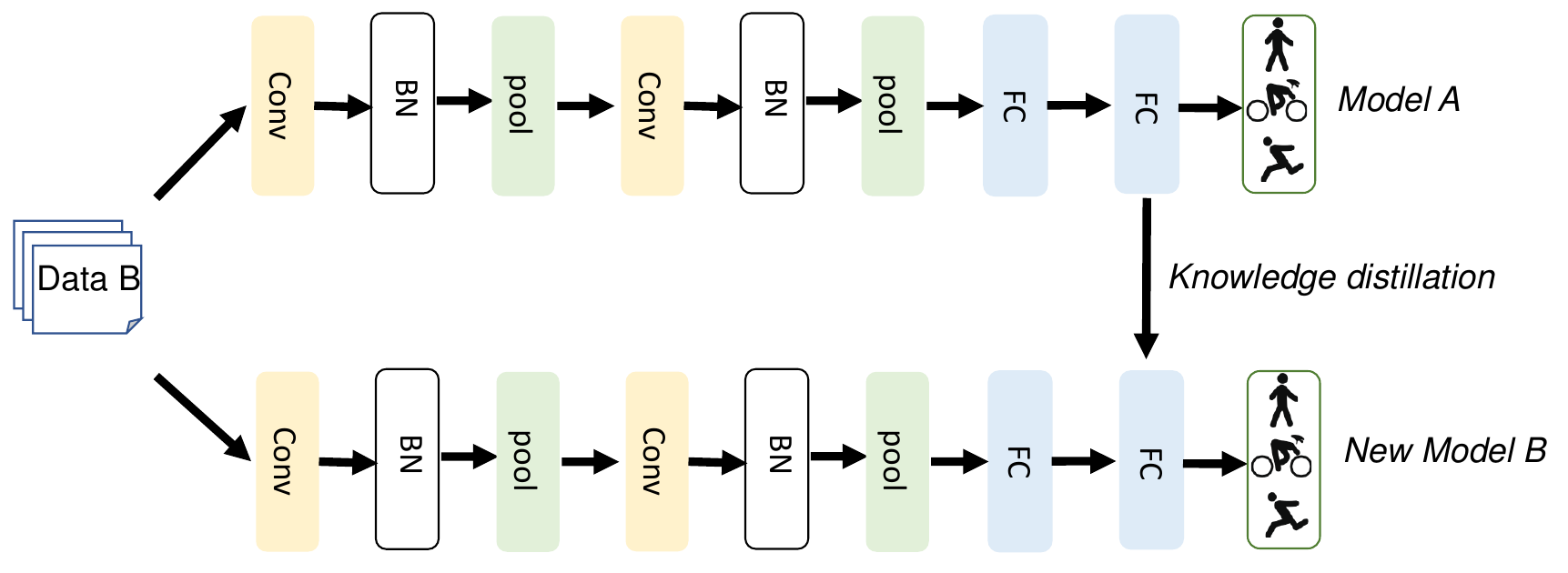}
	\caption{The knowledge transfer between two federations.}
	\label{fig:frame2}
\end{figure}

Without loss of generality, we assume there are four federations, and it can be extended to the more general case easily.
As shown in \figurename~\ref{fig:frame1}, the whole training process is split into two stages, common knowledge accumulation stage (blue arrows) and personalization stage (green arrows). 
In common knowledge accumulation stage, the federations are trained in order and the previous trained one serves as the teacher for the next one. 
The common knowledge accumulation stage lasts for several rounds to ensure each federation's common knowledge are extracted completely. 
Personalization stage is also trained in the same style but the model is sent to the next federation without local training for losing no common knowledge.
From \figurename~\ref{fig:frame1}, we can see no server participates in the training process.
The two stages are both based on feature knowledge distillation (as shown in \figurename~\ref{fig:frame2}),
\begin{equation}
    \ell_{dist}(g_{tea},g_{stu}; \mathbf{x}) = || g_{tea}(\mathbf{x}) - g_{stu}(\mathbf{x}) ||_2^2,
    \label{eqa:fdist}
\end{equation}
where $g_{tea}$ is the feature extractor of the previous federation while $g_{stu}$ is for the current training federation, and $\mathbf{x}$ is a sample of data from the current federation.
Through knowledge distillation, we can make good use of knowledge, viewed as common knowledge, from the previous federation.
Therefore, the total loss to train the local model,$f_i$, is, 
\begin{equation}
    \ell_{total}^i = \frac{1}{n_i^{train}}\sum_{(\mathbf{x},y)\in \mathcal{D}_i^{train}}  \ell_{cls}(f_i;\mathbf{x},y)+\lambda\ell_{dist}(g_{tea}, g_i; \mathbf{x}),
    \label{eqa:total}
\end{equation}
where $\lambda$ is a trade-off of knowledge transfer and focusing the current data while $\ell_{cls}$ is the cross-entropy loss. 
$f_i=c_i\circ g_i$ where $c_i$ is the classification layer and $g_i$ is the feature extractor.
In the following, we will specify the two stages respectively and illustrate how to design $\lambda$ in detail.

\subsection{Common Knowledge Accumulation Stage}
This stage happens in the first part of the whole training process.
In this stage, federations are trained sequentially in a cyclic way and the previous meta knowledge is transferred to the next one via knowledge distillation. 
The knowledge which is useful for the current federation will be preserved via knowledge distillation while the others that are useless will be discarded.
With several rounds of cyclic training, the knowledge that is useful for all federations will be preserved and we denote this type of knowledge as common knowledge. 
Our first stage is just to accumulate common knowledge and the specific detail can be found in \algorithmname~\ref{alg:firststage}.  

\begin{algorithm}[htb]
\caption{Common knowledge accumulation stage}
\label{alg:firststage}
\textbf{Input}: $N$ federations' datasets $\{\mathcal{D}_i\}_{i=1}^N$, $\lambda_0$, $l_{t1}$\\
\textbf{Output}: A common model $f$
\begin{algorithmic}[1] 
\State Initial $\lambda=\lambda_0$
\State Train local models, $f_i$ using $\ell_{cls}$ with $\mathcal{D}_i^{train}$ in each federation
\State Send the current model $f_i$ to the next federation $i+1$
\State Evaluate $f_{i+1}$ on $\mathcal{D}_{i+1}^{valid}$ and obtain $acc_{i+1}^{valid}$
\If{$acc_{i+1}^{valid} > l_{t1}$}
    \State Train $f_{i+1}$ using \equationname~\eqref{eqa:total} with $\mathcal{D}_{i+1}$
\Else
    \State Use $f_i$ to initial $f_{i+1}$
    \State Train $f_{i+1}$ using \equationname~\eqref{eqa:total} with $\mathcal{D}_{i+1}$
\EndIf
\State Repeat steps $3 \sim 10$ until convergence
\State The last model $f_N$ serve as the final common model, $f$
\end{algorithmic}
\end{algorithm}

As shown in \algorithmname~\ref{alg:firststage}, the valid accuracy on the current federation's valid data is used to determine whether to completely preserve the previous federation's knowledge. 
When $acc_{i+1}^{valid} > l_{t1}$, we think the training data on the current federation have enough knowledge to train a model, which means we can discard some unless knowledge via knowledge distillation.
When $acc_{i+1}^{valid} < l_{t1}$, the training data on the current federation has too little information. 
We need to make full use of the previous knowledge and thereby we directly initial the current model with the previous one. 
To preserve the personalization, we may preserve the local BN following FedBN~\cite{li2021fedbn}.
Since we want to accumulate common knowledge in this stage, we fix $\lambda = \lambda_0$ to ensure enough common knowledge preserved. 

\subsection{Personalization Stage}
This stage happens in the second part of the whole training process.
In the above stage, we obtain the common model $f$ which contains enough common knowledge. 
Since no server exists, we have to obtain the personalization models in the same style (sequential) as the first stage. 
To prevent common knowledge loss, we transmit the common $f$ to the next federation before local training. 
The specific detail of the second stage can be found in \algorithmname~\ref{alg:secondstage}. 

\begin{algorithm}[htb]
\caption{Personalization stage}
\label{alg:secondstage}
\textbf{Input}: $N$ federations' datasets $\{\mathcal{D}_i\}_{i=1}^N$, $\lambda_0$, $l_{t2}$, $f$\\
\textbf{Output}: Meta models $\{f_i\}_{i=1}^N$
\begin{algorithmic}[1] 
\State Send the common model $f$ to the next federation $i+1$
\State Evaluate $f$ on $\mathcal{D}_{i+1}^{valid}$ and obtain $acc_{i,i+1}^{valid}$
\State Evaluate $f_{i+1}$ on $\mathcal{D}_{i+1}^{valid}$ and obtain $acc_{i+1,i+1}^{valid}$
\If{$acc_{i, i+1}^{valid} \leq acc_{i+1, i+1}^{valid}$ and $acc_{i, i+1} < l_{t2}$}
    \State Set $\lambda=0$
\Else
    \State Set $\lambda$ via \equationname~\eqref{eqa:person}
\EndIf
\State Train $f_{i+1}$ with \equationname~\eqref{eqa:total}
\State Repeat steps $1 \sim 9$ until all $f_i$ are trained
\end{algorithmic}
\end{algorithm}

When the common model performs seriously terribly on the validation data of the current federation, we want to refer little on it and thereby set $\lambda=0$.
In the first stage, the current $f_i$ has contained other federations' knowledge. 
When the common model's performance is acceptable on the current validation data, we adapt $\lambda$ for personalization:
\begin{equation}
    \lambda= \lambda_0 \times 10^{\min\left( 1,(acc_{i, i+1}^{valid} - acc_{i+1, i+1}^{valid})*5\right)-1}.
    \label{eqa:person}
\end{equation}
Compared to the local model's performance, the better the common model's performance is, the larger $\lambda$ will be.

\subsection{Summary}
We propose a novel framework called MetaFed to facilitate trustworthy FL between different federations.
Here, we re-emphasize these specially designed techniques.
\paragraph{How to ensure enough common knowledge accumulation in clients}
\begin{itemize}
    \item Perform common knowledge accumulation with several rounds.
    \item Transmit enough teacher information according to copy or distillation with fixed $\lambda$.
    \item Incorporate local information
\end{itemize}
\paragraph{How to discard redundant information and obtain personalized models}
\begin{itemize}
    \item Dynamically adjust the amount of inherited information in personalization.
    \item Reduce the influence of other federations' information.
\end{itemize}
\paragraph{How to determine inheritance or discard}
\begin{itemize}
    \item Valid accuracy of each federation can serve as guidance.
\end{itemize}
With these flexible techniques, our framework can ensure personalization without a central server.

\begin{figure}[t!]
	\centering
	\includegraphics[width=0.5\textwidth]{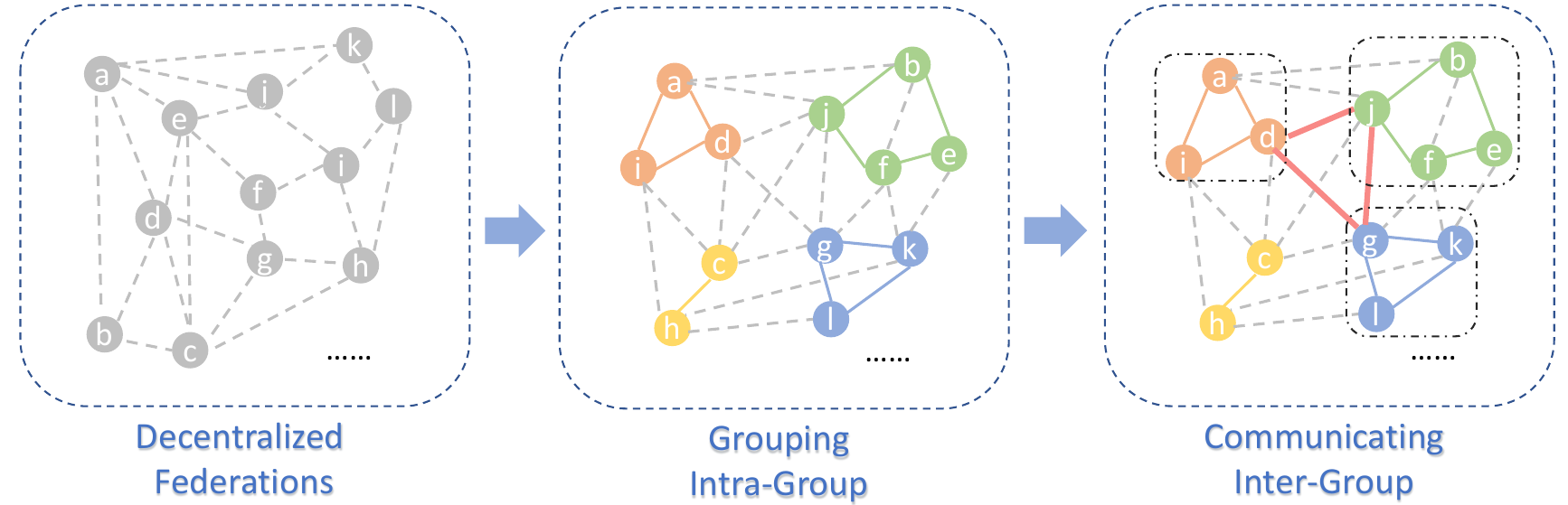}
	\caption{\lw{MetaFed++, a more flexible framework. It first categorizes federations into different groups, and then performs intra-group MetaFed and inter-group MetaFed respectively. Circles Mean federations and lines mean possible links.}}
	\label{fig:frame3}
\end{figure}

\subsection{Extensibility}
\lw{
Surprisingly, MetaFed can be a more flexible framework, just as shown in \figurename~\ref{fig:frame3} and we call this universal situation as MetaFed++.
MetaFed++ imitates the Internet to transmit information.
Federations are scattered around and possible connecting paths are used to transmit information.
MetaFed++ categorizes different federations into multiple groups and performs both inter-group and intra-group communications via cyclic knowledge distillation.
The graph structure is flexible and changeable.
We can even treat groups as new federations and perform MetaFed++ among groups.
Besides, link paths can also be dynamic.
In reality, we can group by geography, organizations, and some other real rules.
In experiments, we can group by similarities among feature statistics following FedAP~\cite{lu2022personalized}. 
Obviously, we can easily extend MetaFed to MetaFed++ for larger and more flexible real-world applications.
}

\subsection{Time Complexity}
\lw{
Denote the number of communication rounds, local train iterations per round, time costs per iteration, and federations as $RI$, $LI$, $TI$, and $N$ respectively.
The time complexity for our method can be $O(RI\times LI \times TI \times N)$ similar to FedAvg. 
However, compared to FedAvg, with limited communication rounds, our method can work better as shown in \figurename~\ref{fig:cost-pf} and \figurename~\ref{fig:cost-label}.
}

\section{Experiments}
\label{sec:exp}
We evaluate the performance of \method on seven benchmarks including time series and image modalities.
The first three benchmarks are on feature shifts where distribution shifts exist in the input space while the other four benchmarks are on label shifts (i.e., distribution shifts exist in the output space).

We compare our method with three state-of-the-art methods including common federated learning methods and some federated learning methods designed for non-iid data, 
\begin{itemize}
    \item FedAvg~\cite{mcmahan2017communication}. Directly aggregate models' parameters without personalization. 
    \item FedProx~\cite{li2018federated}. Allow slight differences between the local model and the global model via a proximal term added to FedAvg.
    \item FedBN~\cite{li2021fedbn}. Preserve the local batch normalization not affected by the other clients.
\end{itemize}
Since these methods all need a server, we ease this restriction for them. 
Adapting these methods without a server will increase communication costs with no performance improvement.
All methods use the same model for fairness.

\subsection{Experiments on Feature Shifts}
\subsubsection{VLCS}
\paragraph{Dataset}
VLCS~\cite{fang2013unbiased} is a popular public image classification dataset.
It comprises four photographic sub-datasets (Caltech101, LabelMe, SUN09, VOC2007) with 10,729 instances of 5 classes. 
Each sub-dataset serves as one federation and there are 4 federations in total. 
Since each dataset contains too many images, we choose $10\%$ for training, $10\%$ for validation, and $20\%$ for testing. 
The validation parts are utilized to guide training and select the best model for each federation.

\paragraph{Implementation Details}
For VLCS, we adopt Alexnet~\cite{krizhevsky2012imagenet} as the feature extractor and a three-layer fully connected neural network as the classifier. 
For model training, we use SGD optimizer with a learning rate of $10^{-2}$.
\lw{For our method, we search best hyperparameters within $[0.1,1,5,10]$ for $\lambda_0$ and $[0,0.4,0.5,0.6,1.1]$ for $l_{t1}$.
More analysis on hyperparameter sensitivity can be found in \paragraphname~\ref{psec:paras}.}
All methods are implemented in the same environment for fairness and we run three trials to record the average accuracy.

\paragraph{Results}
\input{tab-vlcs}

The classification results for each federation on VLCS are shown in \tablename~\ref{tab:my-table-vlcs}.
We have the following observations from these results.
1) Our method achieves the best effects on average with a remarkable improvement (over $4.25\%$ compared to FedBN). 
We even almost achieve the best performance on each federation which demonstrates the superiority and the capability of personalization of our method.
2) Since it is a feature shift situation, FedBN has a better performance compared to FedAvg. FedProx has an acceptable performance.

\subsubsection{PACS}
\paragraph{Dataset}
PACS~\cite{li2017deeper} is another popular object classification benchmark.
It also consists of four sub-datasets (photo, art-painting, cartoon, sketch) with 9991 images of seven classes.
There exist large discrepancies in image styles among different sub-datasets.
Similar to VLCS, we choose $10\%$ for training, $10\%$ for validation, and $20\%$ for testing. 
We also utilize validation parts for the guidance and selections.

\paragraph{Implementation Details}
We utilize the same architecture and the same optimizer as VLCS.
All methods are implemented in the same environment for fairness and we run three trials to record the average accuracy.

\paragraph{Results}
The classification results for each federation on PACS are shown in \tablename~\ref{tab:my-table-pacs}.
We have the following observations from these results.
1) Without a doubt, our method achieves the best performance on each federation. On average, it has a significant improvement with $4.23\%$ compared to FedBN. 
2) Similar to VLCS, in this feature shift situation, FedBN has a better performance compared FedAvg and FedProx.
Different from VLCS, FedAvg and FedProx are far worst than FedBN, which can be due to larger divergences, more classes, etc.

\input{tab-pacs}

\subsubsection{Human Activity Recognition}
\paragraph{Dataset}
To further prove the superiority of our method, we also adopt a public time-series benchmark with feature shifts, PAMAP2~\cite{reiss2012introducing}.
PAMAP2 is a human activity recognition with data of 18 human activities performed by 9 subjects.
We use 3 inertial measurement units' data with 27 channels and utilize the sliding window technique to preprocess data. 
12 classes with 12291 instances are selected. 
Since there are no natural sub-datasets in PAMAP2, we split artificially PAMAP2 into four sub-datasets according to persons and we call this task a cross-person task~\cite{lu2022domain}. 
9 subjects are divided into four groups, [3,2,8],[1,5],[0,7],[4,6] where different numbers denote different persons, and we try our best to make each domain have a similar number of data.
Similar to VLCS, we choose $10\%$ for training, $10\%$ for validation, and $20\%$ for testing. 
We also utilize validation parts for the guidance and selections.

\paragraph{Implementation Details}
For PAMAP2, we utilize a CNN composed of two convolutional layers, two pooling layers, two batch normalization layers, and two fully connected layers~\cite{wang2019deep}. 
Other settings are similar to VLCS.

\paragraph{Results}

The classification results for each federation on PAMAP2 in the Cross-Person setting are shown in \tablename~\ref{tab:my-table-xp}.
We have the following observations from these results.
1) Our method achieves the best performance on average with an improvement of $1.04\%$. 
However, it is slightly worse than FedBN in Federation 0 and slightly worse than FedProx in Federation 3.
We think that the task is so simple that some methods, e.g. FedProx, with few techniques for feature shifts can work slightly better.
And the improvements on some federations with a drop less than $1\%$ do not illustrate that our method is unacceptable.
2) Similar to VLCS, in this feature shift situation, FedBN has a better performance compared FedAvg and FedProx.
Simultaneously, FedAvg and FedProx have a similar performance to FedBN.

\input{tab-xp}

\subsubsection{Summary}

In the feature shift situation, whatever data are visual images or time series, our method achieves the best performance compared to other methods.
The more difficult the task is, the larger improvements our method has.
Moreover, FedBN, a method specifically designed for the feature shift, has the second-best performance.

\subsection{Experiments on Label Shifts}
\subsubsection{Human Activity Recognition}

\begin{figure}[ht!]
	\centering
	\subfigure[PAMAP2]{
		\label{fig:datadis-p}
		\includegraphics[height=0.16\textwidth]{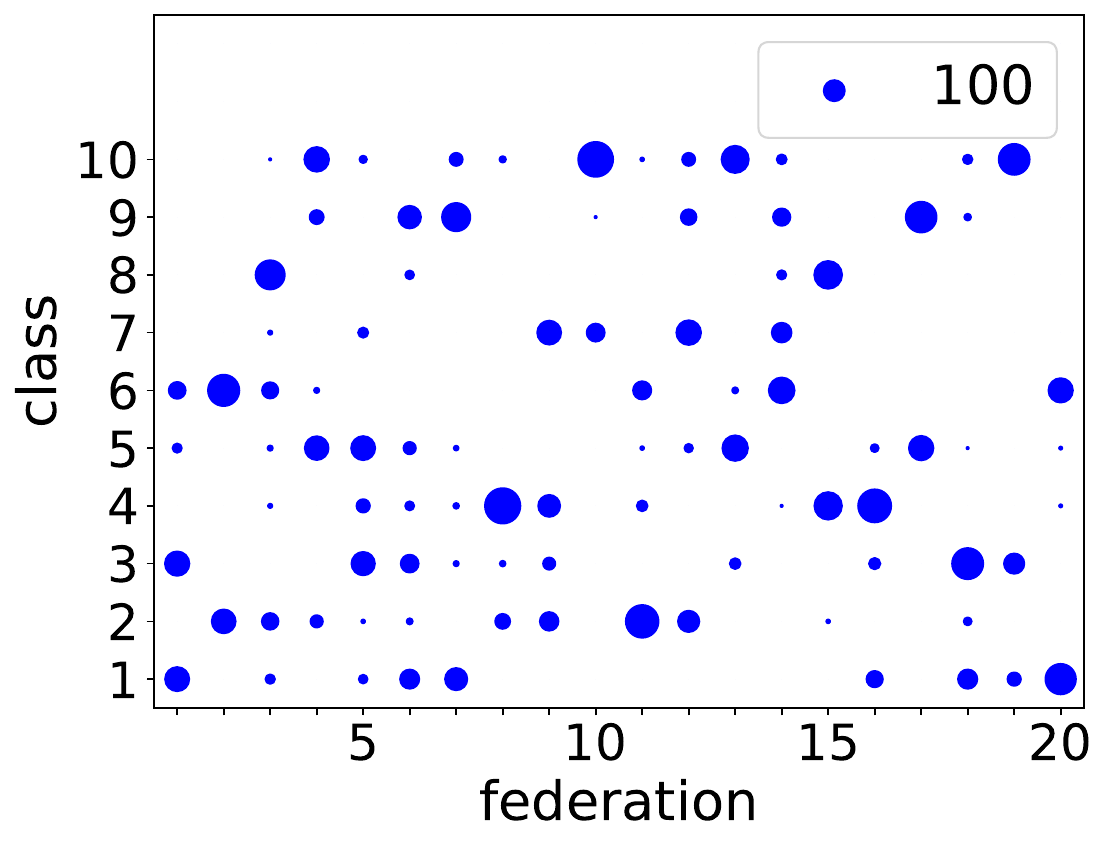}
	}
	\subfigure[OrganA]{
		\label{fig:datadis-a}
		\includegraphics[height=0.16\textwidth]{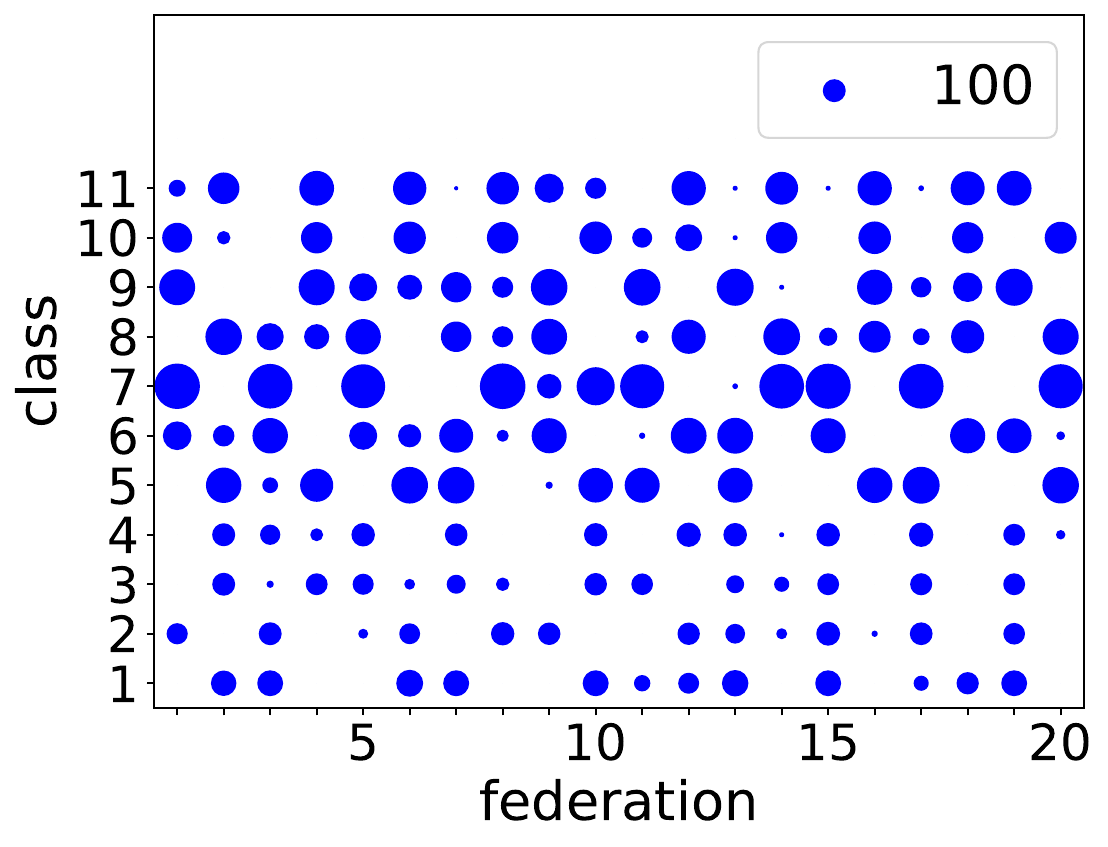}
	}
	\subfigure[OrganC]{
		\label{fig:datadis-c}
		\includegraphics[height=0.16\textwidth]{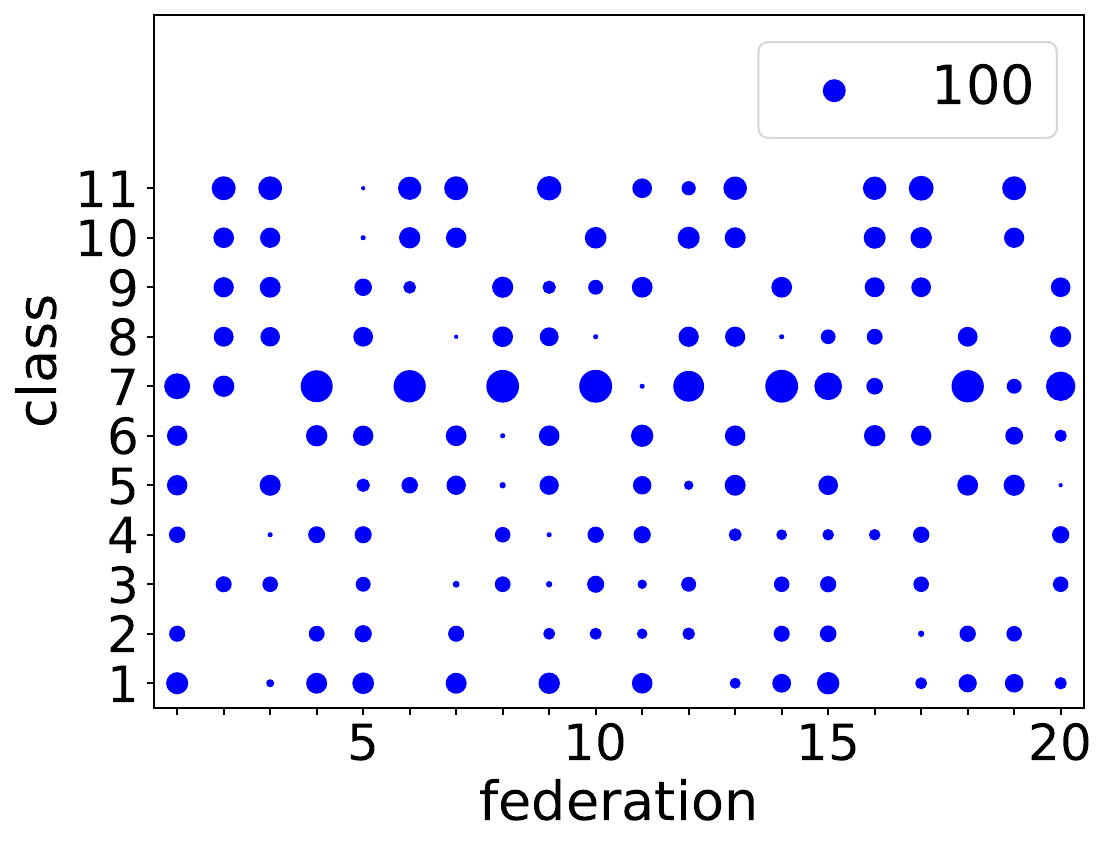}
	}
	\subfigure[OrganS]{
		\label{fig:datadis-s}
		\includegraphics[height=0.16\textwidth]{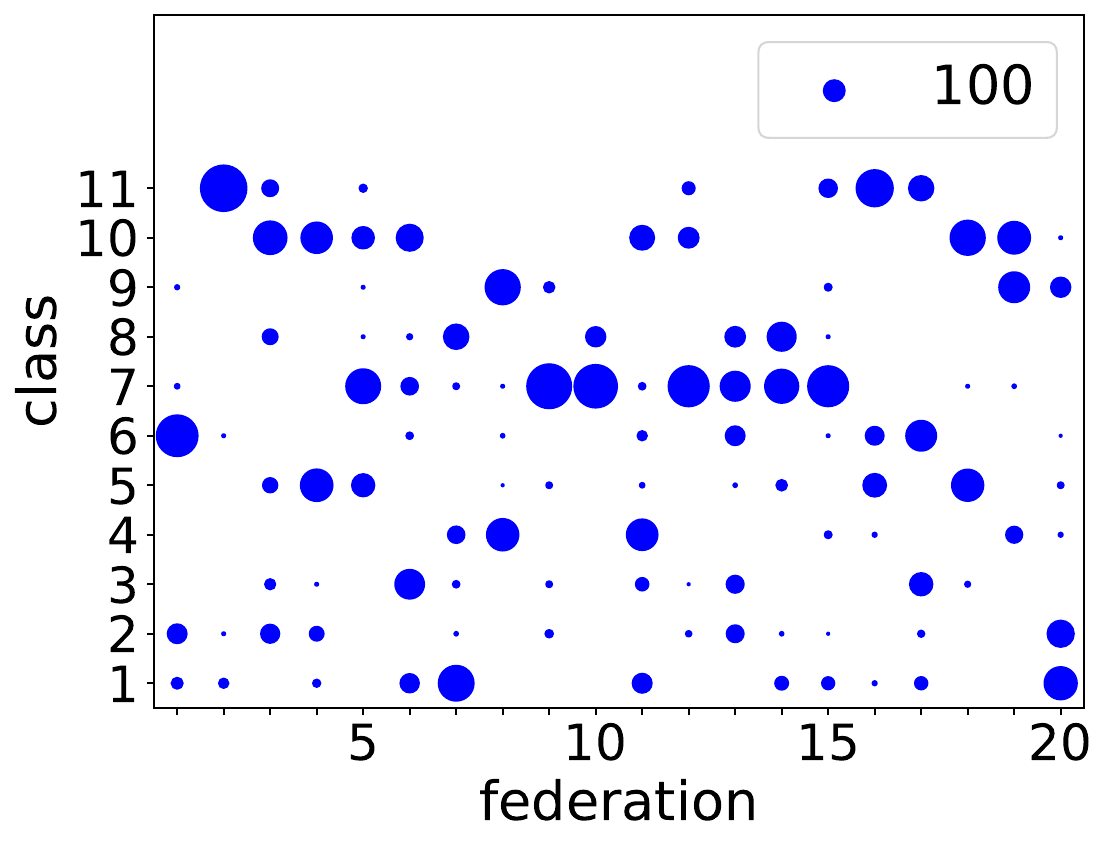}
	}
	\caption{\lw{The number of samples per class allocated to each federation (indicated by dot size).}}
	\label{fig:datasplit}
\end{figure}

\paragraph{Dataset} 
For the label shift benchmarks, we first adopt a public time-series benchmark, PAMAP2~\cite{reiss2012introducing}. 
PAMAP2 contains data of 18 human activities performed by 9 subjects. 
We also use 3 inertial measurement units' data with 27 channels and utilize the sliding window technique to preprocess data. 
10 classes with 17639 instances are selected. 
To simulate labels shift, we follow \cite{yurochkin2019bayesian} and use Dirichlet distributions to create disjoint non-iid. training data. 
\figurename~\ref{fig:datadis-p} visualized how samples are distributed. 
For each federation, $40\%, 30\%,$ and $30\%$ of data are used for training, validation, and testing respectively.
\input{tab-pamap-tn}
\input{tab-medmnist-tn}

\paragraph{Implementation Details} 
For PAMAP2, we utilize a CNN composed of two convolutional layers, two pooling layers, two batch normalization layers, and two fully connected layers~\cite{wang2019deep}. 
Other settings are similar to VLCS.

\paragraph{Results}
The classification results for each federation on PAMAP2 are shown in \tablename~\ref{tab:my-table-pamap}.
We have the following observations from these results.
1) Our method also achieves the best effects on average with a remarkable improvement (over $2.63\%$ compared to FedBN) in this situation where label shifts exist. 
2) In this situation, 20 federations in total make the problem more complicated.
Although our method does not achieve the best performance on each federation, it still achieves acceptable results on almost every federation. 
3) Compared to FedAvg and FedProx, FedBN and our method achieve remarkable improvement, which may illustrate that FedBN can also cope with label shifts sometimes.

\subsubsection{Medical Image Classification}
\paragraph{Dataset} 
To further validate our method, we evaluate our method on three public medical image classification benchmarks. 
We choose 3 datasets, OrganAMNIST, OrganCMNIST, and OrganSMNIST~\cite{bilic2019liver,xu2019efficient}, from a larger-scale MNIST-like collection of standardized biomedical images, MedMNIST~\cite{medmnistv1,medmnistv2}. 
These three datasets are all about Abdominal CT images with 11 classes and they have 58,850, 23,660, and 25,221 samples respectively.
Similar to PAMAP2, we utilize Dirichlet distribution to split data, and \figurename~\ref{fig:datadis-a}-\figurename~\ref{fig:datadis-s} visualize how samples are distributed. 
In each federation, $40\%, 30\%,$ and $30\%$ of data are used for training, validation, and testing respectively.

\begin{figure}[ht!]
	\centering
	\subfigure[Average Acc]{
		\label{fig:abl-a-pf}
		\includegraphics[height=0.15\textwidth]{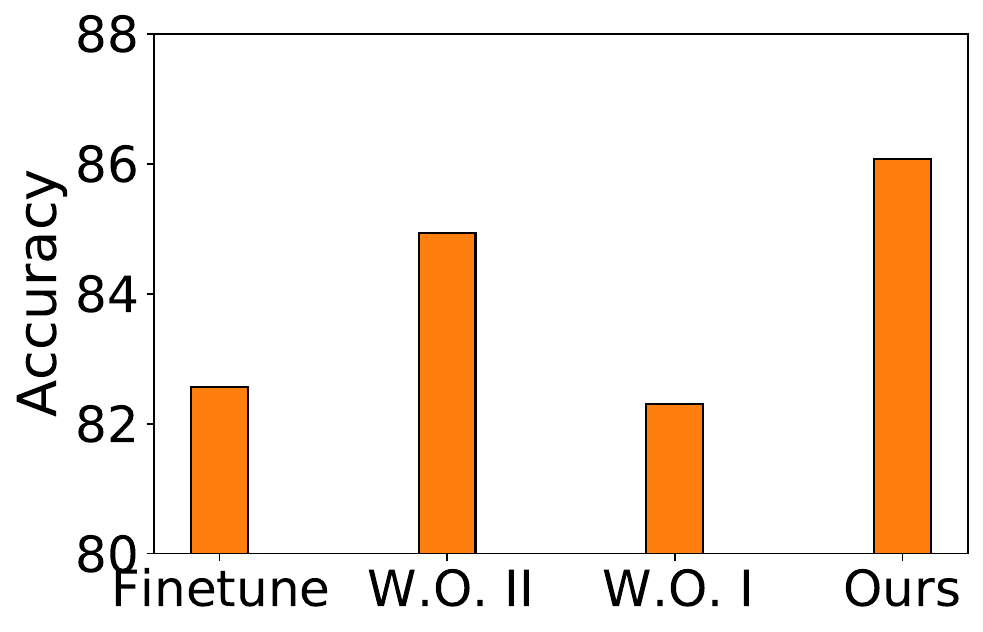}
	}
	\subfigure[Federation Acc]{
		\label{fig:abl-s-pf}
		\includegraphics[height=0.15\textwidth]{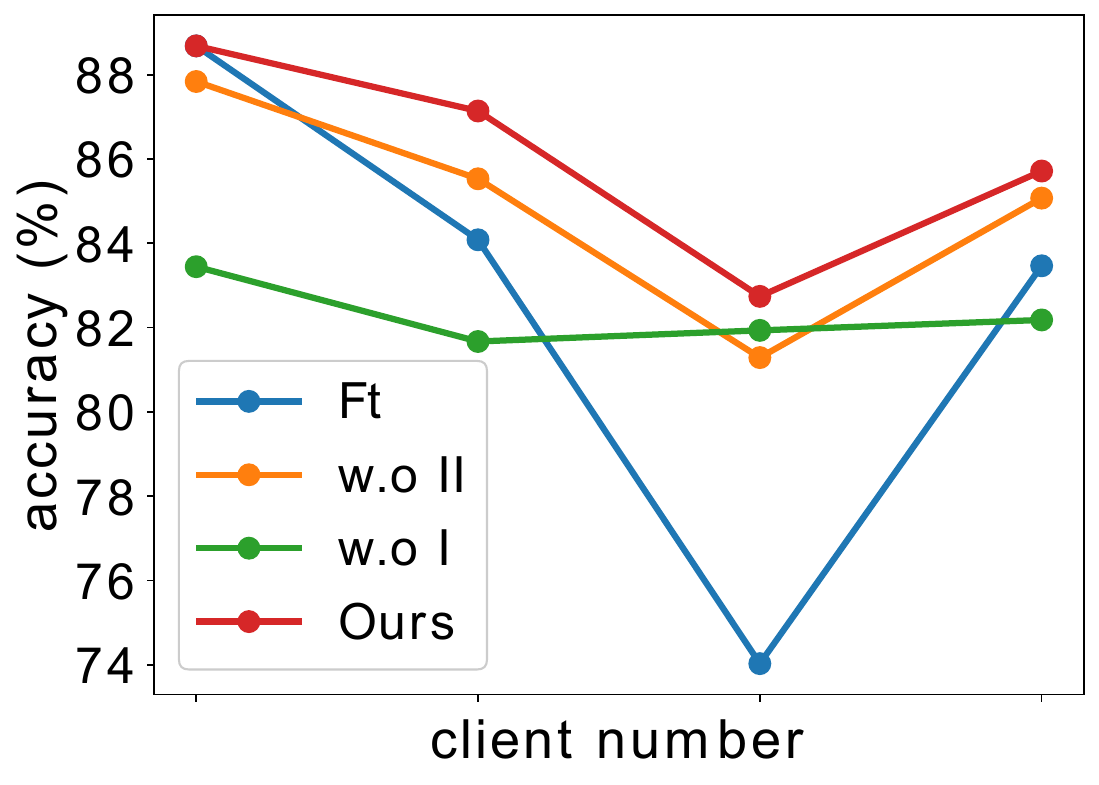}
	}
	\caption{Ablation study on PAMAP2 under feature shifts.}
	\label{fig:abla-pf}
\end{figure} 

\begin{figure}[ht!]
	\centering
	\subfigure[Average Acc]{
		\label{fig:abl-a-p}
		\includegraphics[height=0.15\textwidth]{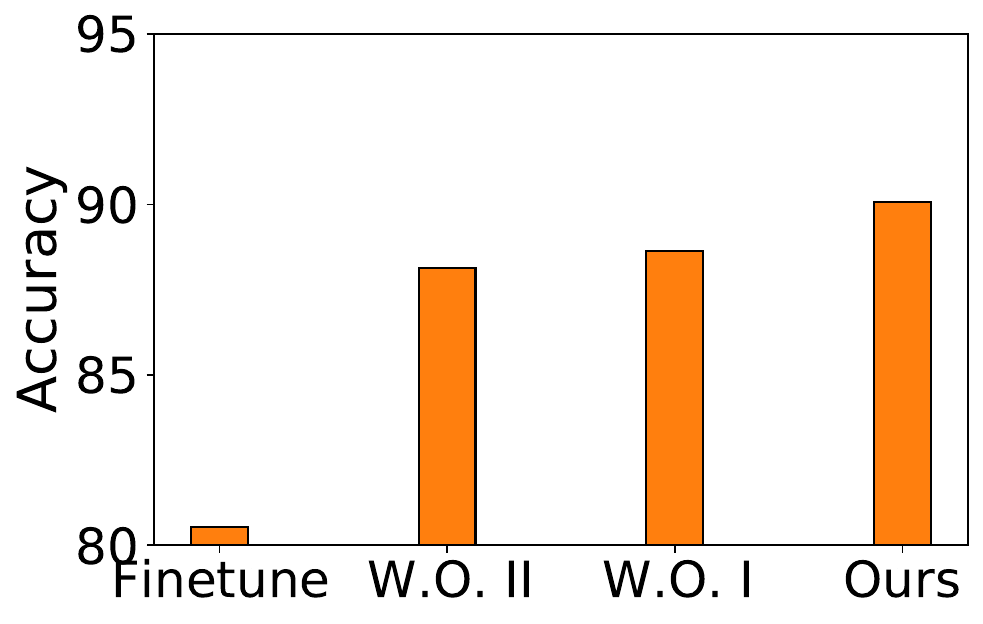}
	}
	\subfigure[Federation Acc]{
		\label{fig:abl-s-p}
		\includegraphics[height=0.15\textwidth]{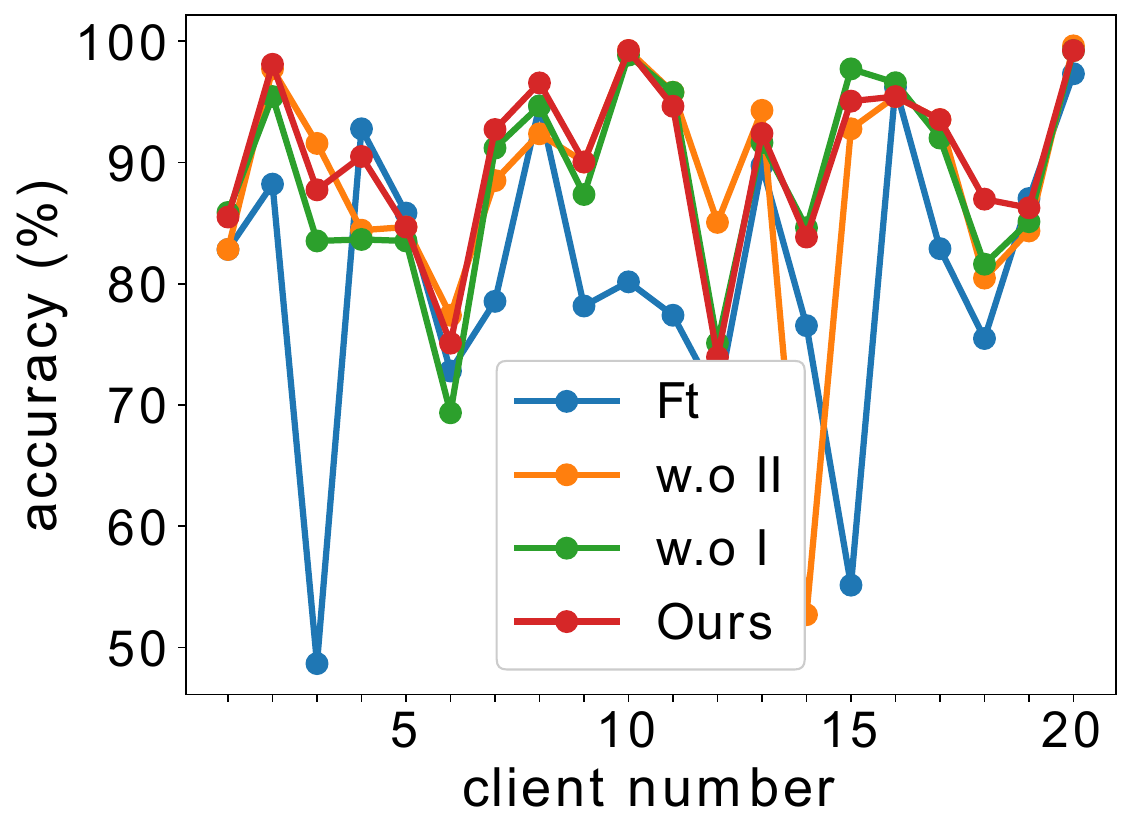}
	}
	\caption{Ablation study on PAMAP2 under label shifts.}
	\label{fig:abla-p}
\end{figure} 

\begin{figure*}[ht!]
	\centering
	\subfigure[$\lambda_0$ (Label shift)]{
		\label{fig:sens-l-p}
		\includegraphics[height=0.1\textwidth]{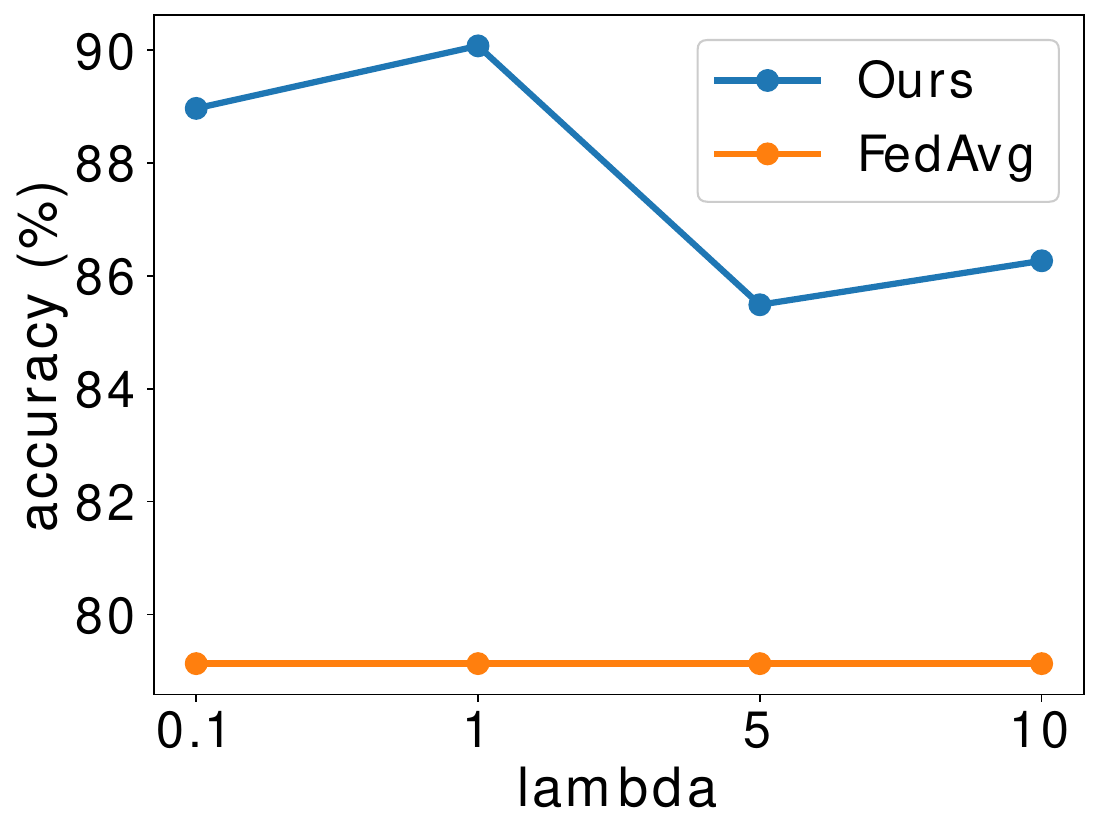}
	}
	\subfigure[$l_{t1}$ (Label shift)]{
		\label{fig:sens-t-p}
		\includegraphics[height=0.1\textwidth]{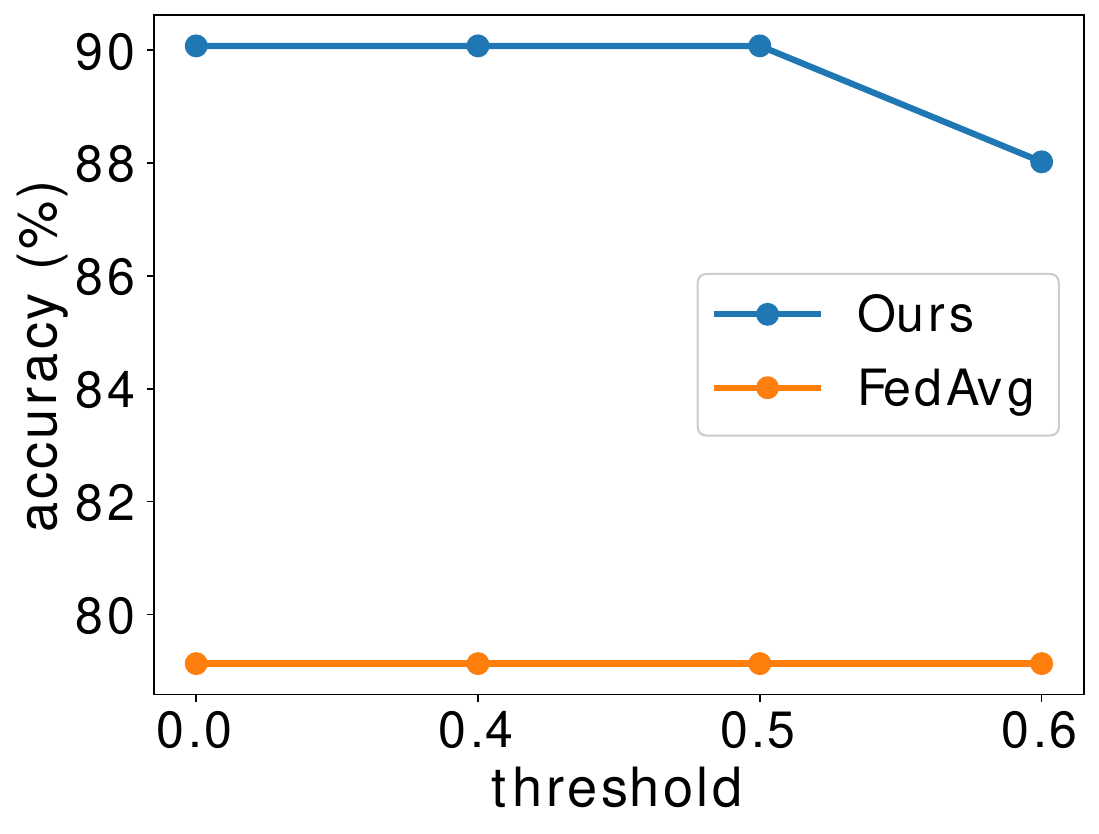}
	}
    \subfigure[Distillation Style (Label shift)]{
		\label{fig:sens-dis-p}
		\includegraphics[height=0.1\textwidth]{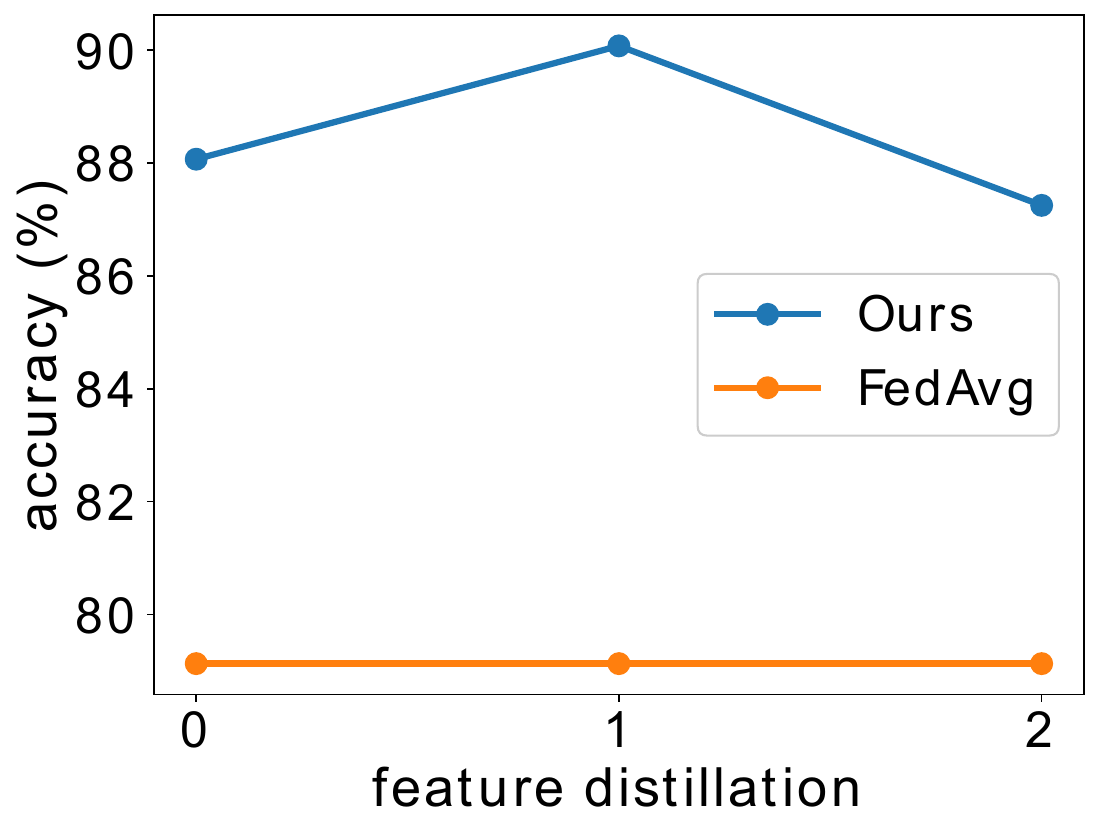}
	}
  \subfigure[Sort (Label shift)]{
		\label{fig:sens-sort-p}
		\includegraphics[height=0.1\textwidth]{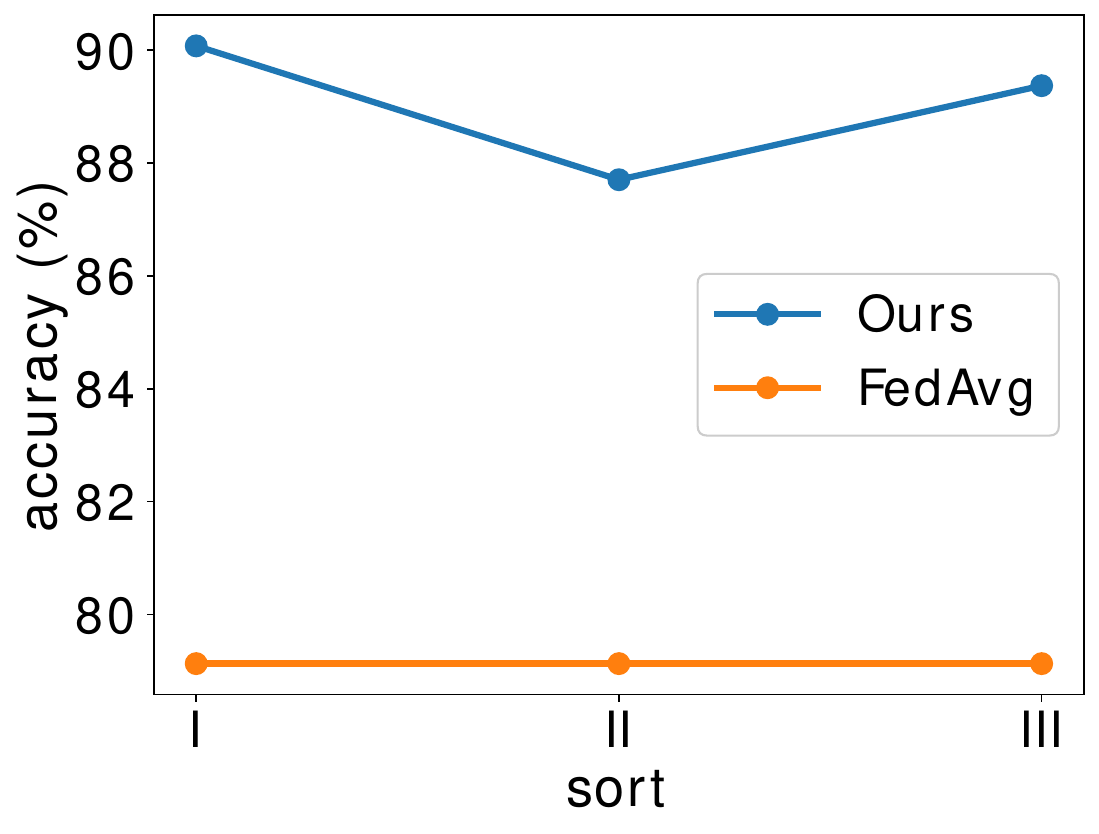}
	}
	\subfigure[ShareBN (Label shift)]{
		\label{fig:sens-bn-p}
		\includegraphics[height=0.1\textwidth]{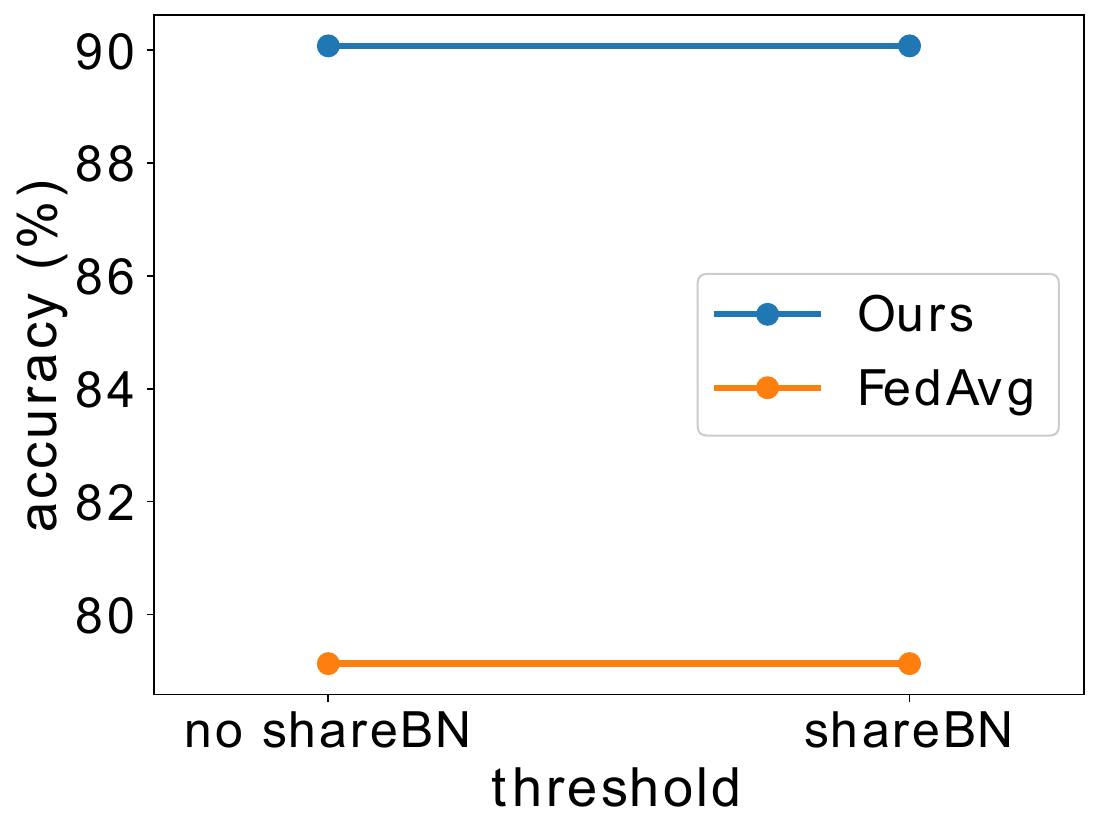}
	}
 \subfigure[ShareBN (Feature shift)]{
		\label{fig:sens-bn-pf}
		\includegraphics[height=0.1\textwidth]{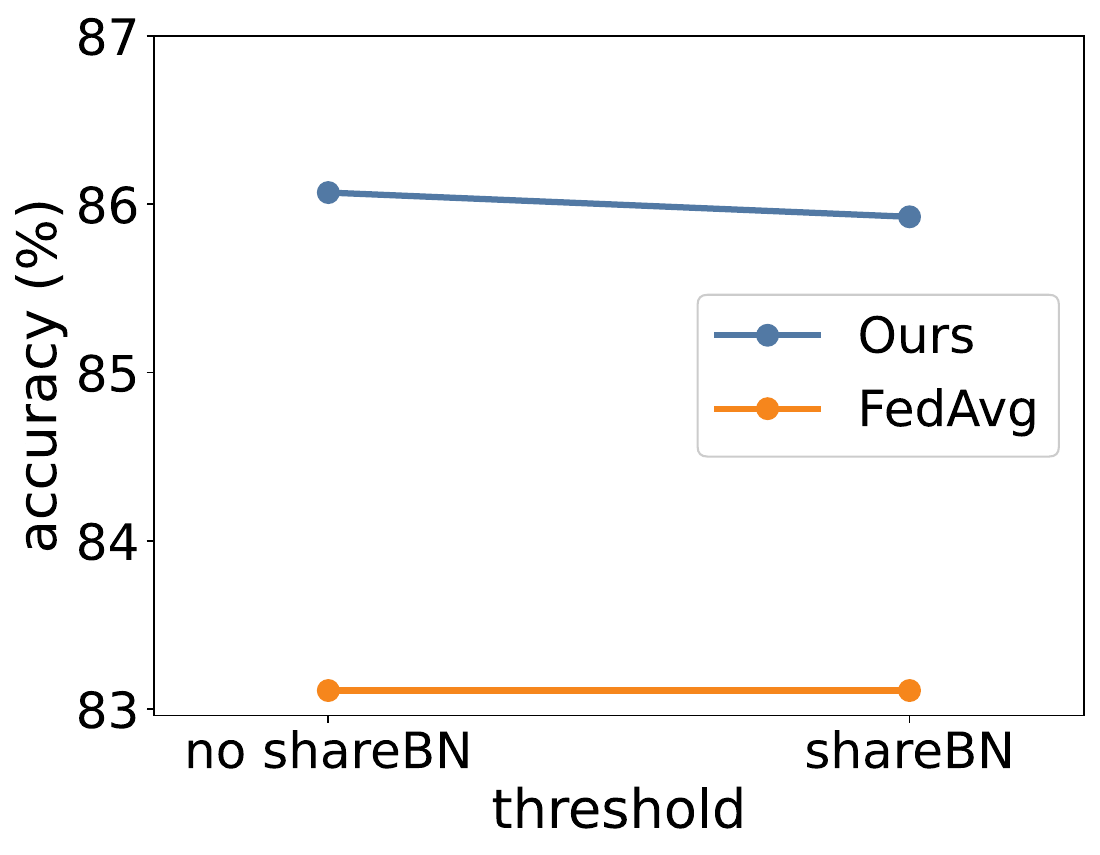}
	}
	\caption{Parameter sensitivity on PAMAP2 under label shifts and feature shifts. \figurename~\ref{fig:sens-l-p}-\figurename~\ref{fig:sens-sort-p} are under label shifts while \figurename~\ref{fig:sens-bn-pf} is under feature shifts.}
	\label{fig:sens-p}
\end{figure*} 

\begin{figure*}[ht!]
	\centering
	\vspace{-.2in}
    \subfigure[Feature shift]{
		\label{fig:cost-pf}
		\includegraphics[height=0.12\textwidth]{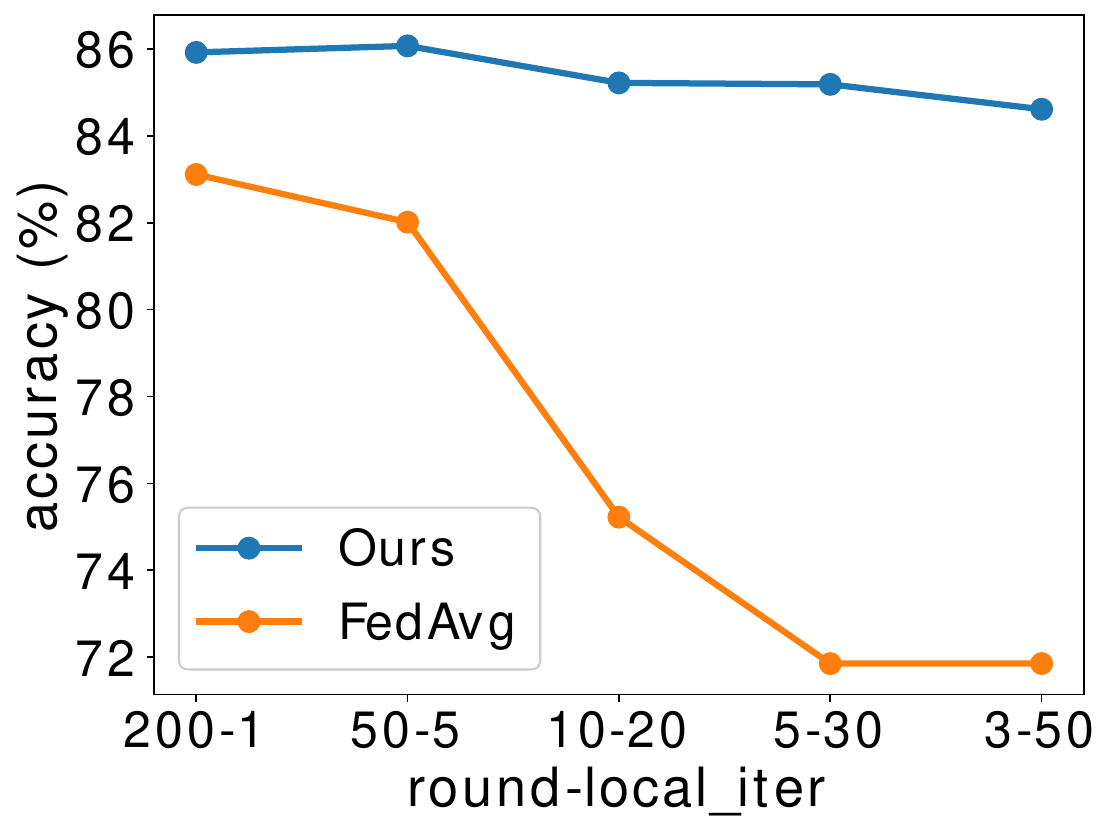}
	}
	\subfigure[Label shift]{
		\label{fig:cost-label}
		\includegraphics[height=0.12\textwidth]{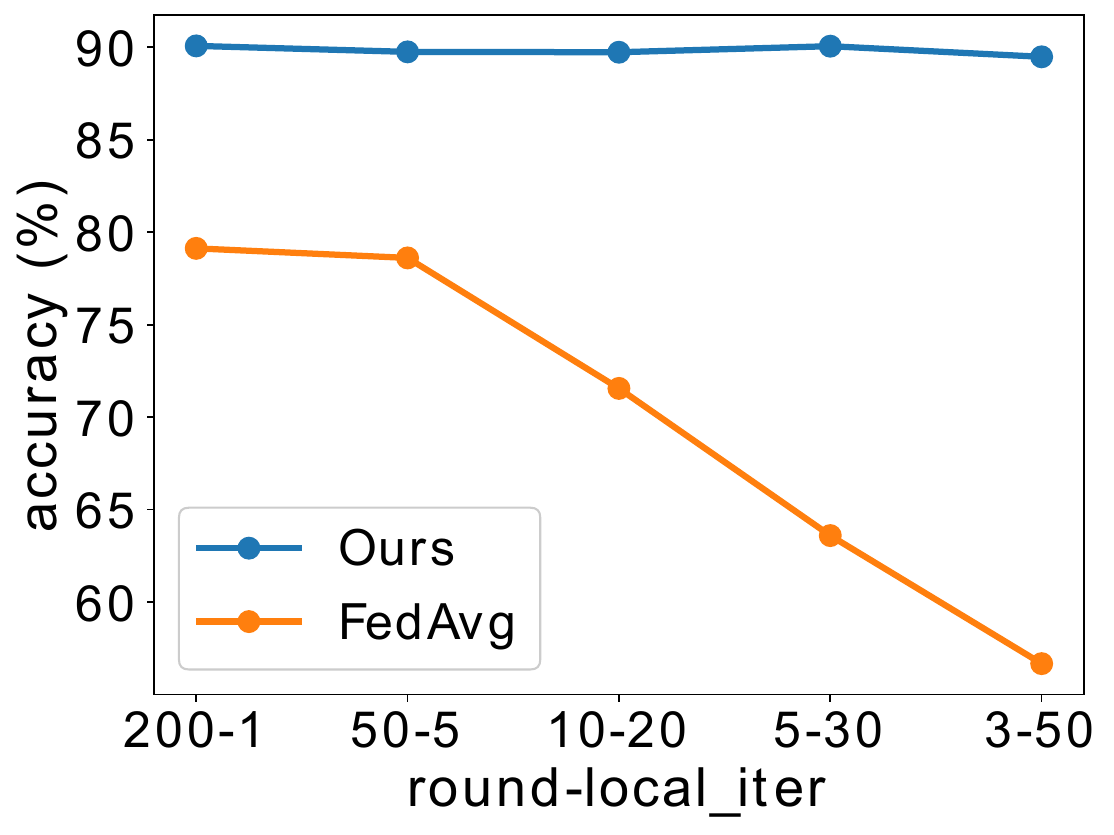}
	}
	\subfigure[Average Acc]{
		\label{fig:avc-++}
		\includegraphics[height=0.12\textwidth]{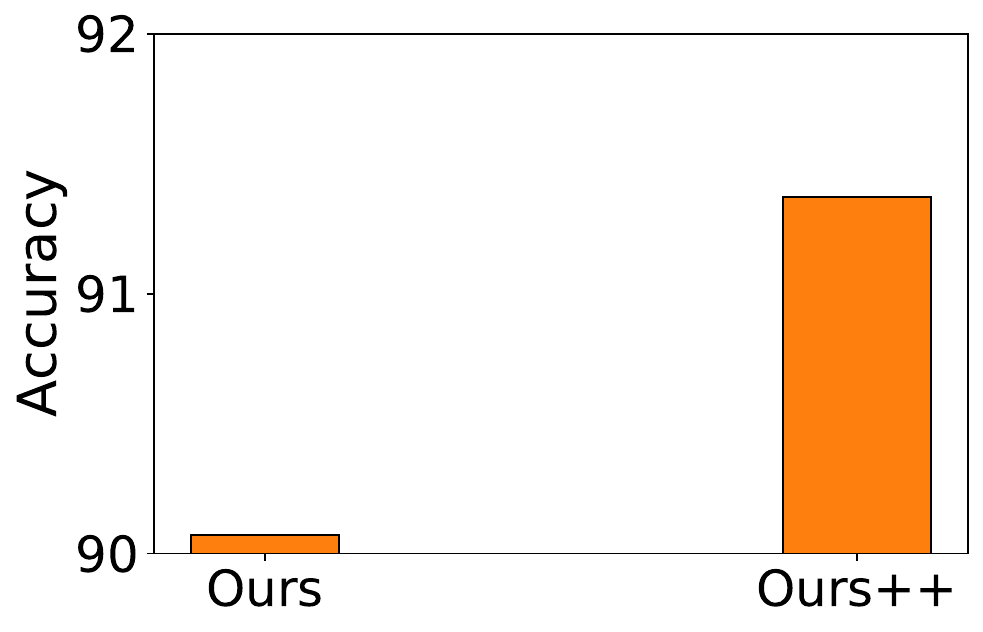}
	}
	\subfigure[Federation Acc]{
		\label{fig:fvc-++}
		\includegraphics[height=0.12\textwidth]{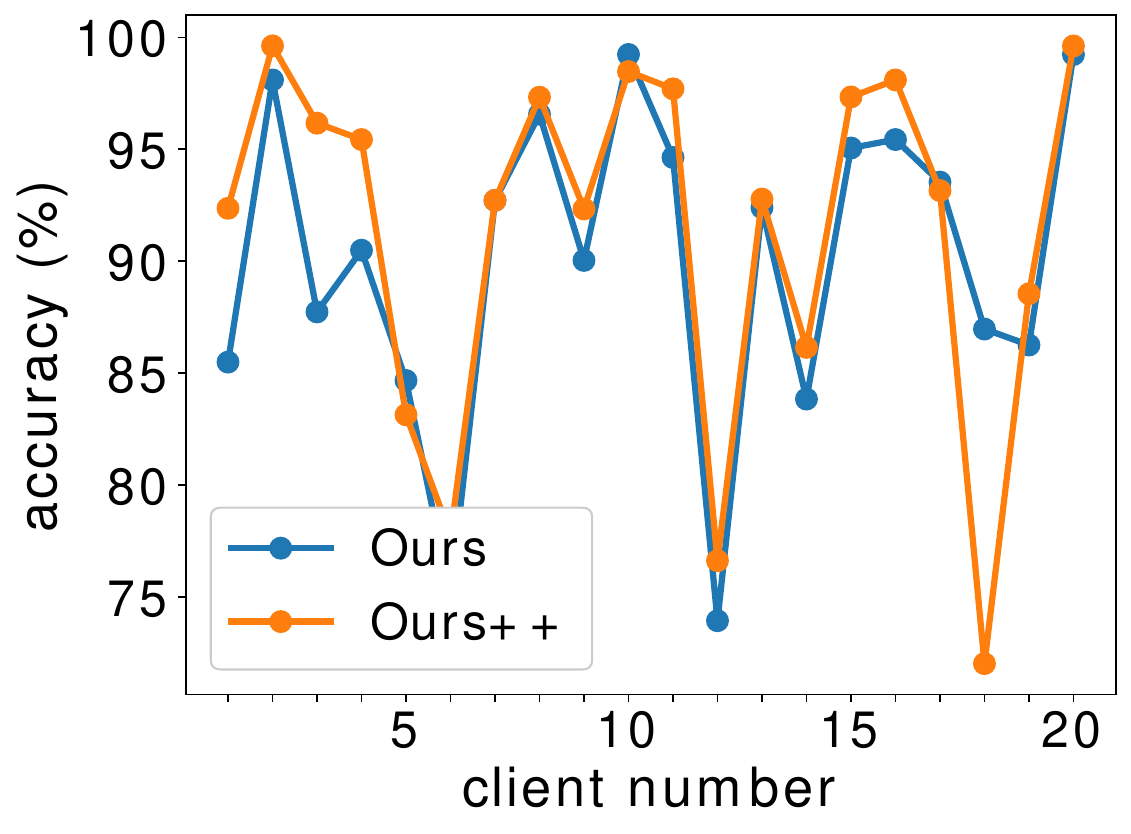}
	}
	\subfigure[Convergence]{
		\label{fig:conver}
		\includegraphics[height=0.12\textwidth]{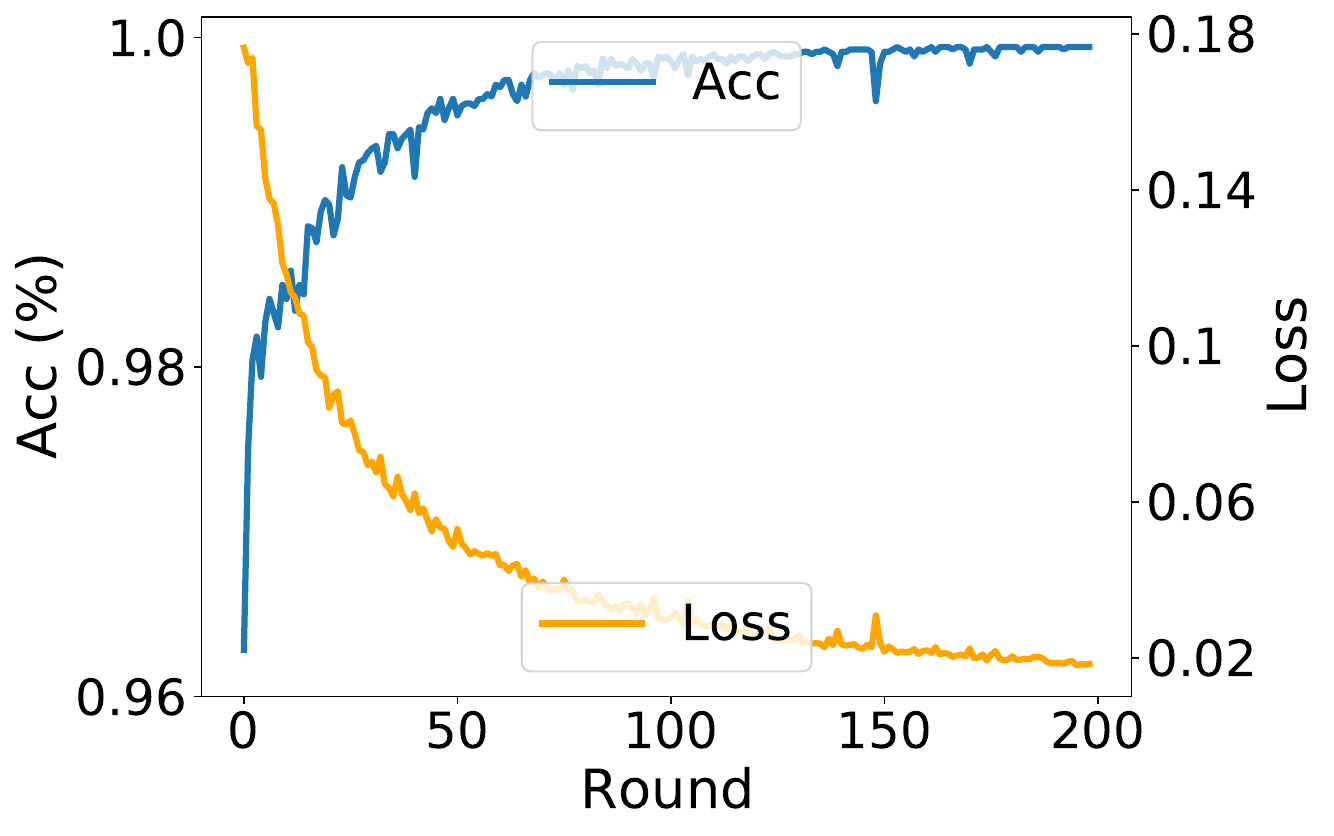}
	}
	\caption{ Analysis on PAMAP2. \figurename~\ref{fig:cost-pf}-\figurename~\ref{fig:cost-label} are on Performance w. communication. \figurename~\ref{fig:avc-++}-\figurename~\ref{fig:fvc-++} are on performance of MetaFed++. \figurename~\ref{fig:conver} is on convergence.}
    \label{fig:analysis}
    	\vspace{-.2in}
\end{figure*} 

\paragraph{Implementation Details}
For these three datasets, we utilize adapted LeNet5~\cite{lecun1998gradient} due to the image size with $28 \times 28$. 
Other settings are similar to VLCS.

\paragraph{Results}
The classification results for each federation on OrganAMNIST, OrganCMNIST, and OrganSMNIST, are shown in \tablename~\ref{tab:my-table-medmnist}. 
We have the following observations from these results.
1) Our method also achieves the best effects on average with a remarkable improvement (over $3.83\%$, $3.00\%$, and $12.91\%$ respectively) in this situation where label shifts exist. 
2) When federation distributions have small differences (\figurename~\ref{fig:datadis-a} and \figurename~\ref{fig:datadis-c}), three state-of-the-art methods have similar performance and ours achieves remarkable improvements.
When federations have huge differences from each other (\figurename~\ref{fig:datadis-s}), FedBN can achieve a remarkable improvement compared to FedAvg and FedProx while ours shows another crazy improvement compared to FedBN
3) The above experiments demonstrate that our method can achieve the best performance in both two settings.

\subsubsection{Summary}

In the label shift situation, whatever data are visual images or time series, our method also achieves the best performance compared to other methods.
Interestingly, FedBN, a method specifically designed for the feature shift, still has the second-best performance.

\subsection{Analysis}

\paragraph{Ablation Study}
We also perform ablation study to illustrate the effects of each part of our methods. 
As shown in \figurename~\ref{fig:abla-pf} and \figurename~\ref{fig:abla-p}, we can see that replacing knowledge distillation with fine-tuning (Finetune), training without common knowledge accumulation stage (W.O. I), and training without personalization stage (W.O. II) will all make performance drop, which demonstrates that each part of our method can all bring benefits.
We do not use the common $f$ for testing but each local model with local adaptation brought by knowledge distillation does testing.
There also exist slight differences between results under feature shifts and label shifts.
When feature shifts exist, common knowledge accumulation stage may play a more important role compared to personalization stage and there is a further improvement with personalization stage.
Finetune seems to bring no improvement.
When label shifts exist, Fine-tune can bring slight improvements while both two stages work well.

\paragraph{Parameter Sensitivity}
\label{psec:paras}

	

In this part, we evaluate five hyperparameters, $\lambda_0$, $l_{t1}$, feature distillation styles\footnote{We utilize three styles to compute feature distillation loss. They utilize features of the output of the last convolutional block, the outputs of the penultimate layer, and the combinational outputs of last few layers.}, whether to share BN parameters, and the transmission sorts, in \method.
We change one parameter and fix the others.
From \figurename~\ref{fig:sens-l-p}-\figurename~\ref{fig:sens-sort-p}, we can see that our method can always achieve better performance than FedAvg under label shifts whatever each hyperparameter is.
Among five hyperparameters, whether to share BN parameters has minimal impact on the final results.
We also perform parameter sensitivity on PAMAP2 under feature shifts.
As shown in \figurename~\ref{fig:sens-bn-pf}, our method still achieves better performance compared to FedAvg under feature shifts.
Moreover, similar to the results under label shifts, whether to share BN parameters has slight impact on the final results.
The results reveal that \method is more effective and robust than other methods under different hyperparameters in most cases.

\paragraph{Communication Costs}


To further prove that our method can reduce communication costs, we increase the local training iteration number and decrease the total communication rounds to evaluate our method and the baseline. 
As shown in \figurename~\ref{fig:cost-pf}, when communication costs are limited under feature shifts, our method has related stable results while FedAvg drops seriously. 
A similar phenomenon exists in the situation under label shifts as shown in \figurename~\ref{fig:cost-label}.
In realistic applications, the communication cost is an important evaluation metric and there are often strictly limited costs. 
Therefore, few communication costs with stable and acceptable performance are vital.


\paragraph{MetaFed++}

\lw{
We evaluate MetaFed++ in this part.
We group 20 federations into 3 groups and perform MetaFed both inter-group and intra-group.
As shown in \figurename~\ref{fig:avc-++} and \figurename~\ref{fig:fvc-++}, MetaFed++ even achieves a better average accuracy compared to MetaFed, which means our method can be easily deployed in real large applications. 
In addition, although MetaFed++ performs worse than MetaFed in some federations, MetaFed++ performs better than MetaFed in most federations.
This can be due to that federations are more likely to be influenced by the federations within the same groups.
}

\paragraph{Convergence}
\lw{
We offer the convergence analysis in this part.
As shown in \figurename~\ref{fig:conver}, both the average loss and the average accuracy are convergent and they end in stable and ideal states, which demonstrates the superiority of \method.
}

\paragraph{Compare to More Methods}

\begin{table}[htbp]
\centering
\caption{\lw{Comparison to BrainTorrent on PAMAP2.}}
\resizebox{0.3\textwidth}{!}{
\begin{tabular}{ccc}
\toprule
 & BrainTorrent          & \method \\ \midrule
Feature shifts&80.59&86.07\\
Label shifts&83.55&90.07\\
 \bottomrule
\end{tabular}}

\label{tab:my-table-brcom}
\end{table}
\lw{
In this part, we provide some comparisons to BrainTorrent~\cite{roy2019braintorrent}, a peer-to-peer method.
As shown in \tablename~\ref{tab:my-table-brcom}, our method still achieves better performance under both feature shifts and label shifts.
Besides, our method saves storage space compared to BrainTorrent.
}

\section{Real-world Applications}
\label{sec-realword}
\subsection{COVID-19}
\subsubsection{Background}
\begin{figure}[ht!]
	\centering
	\subfigure[Normal]{
		\label{fig:c19-n}
		\includegraphics[height=0.13\textwidth]{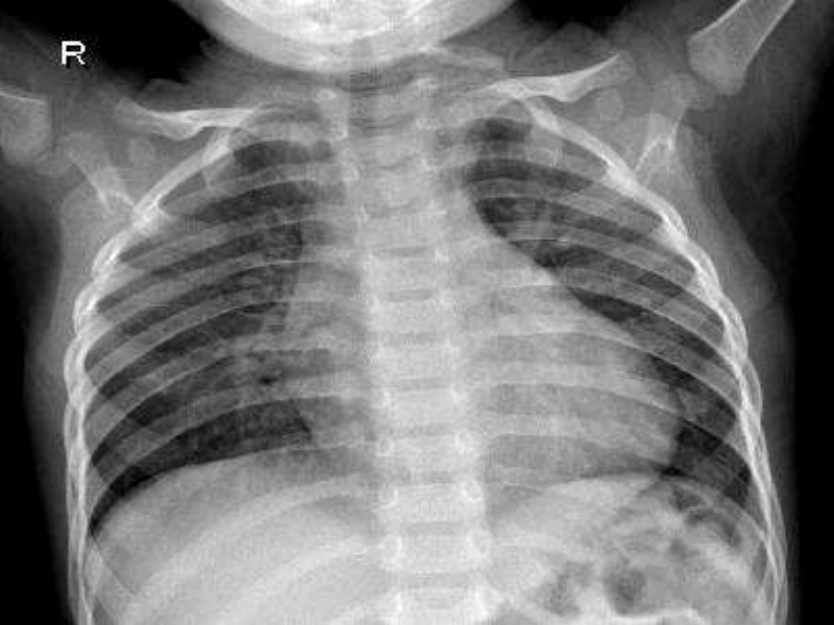}
	}
	\subfigure[Pneumonia-Bacterial]{
		\label{fig:c19-pb}
		\includegraphics[height=0.13\textwidth]{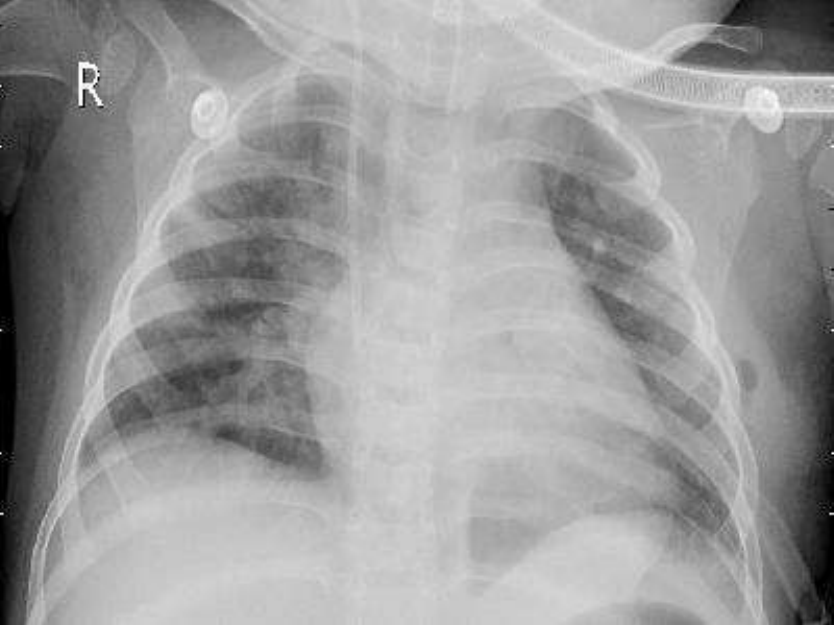}
	}
    \subfigure[Pneumonia-Viral]{
		\label{fig:c19-pv}
		\includegraphics[height=0.13\textwidth]{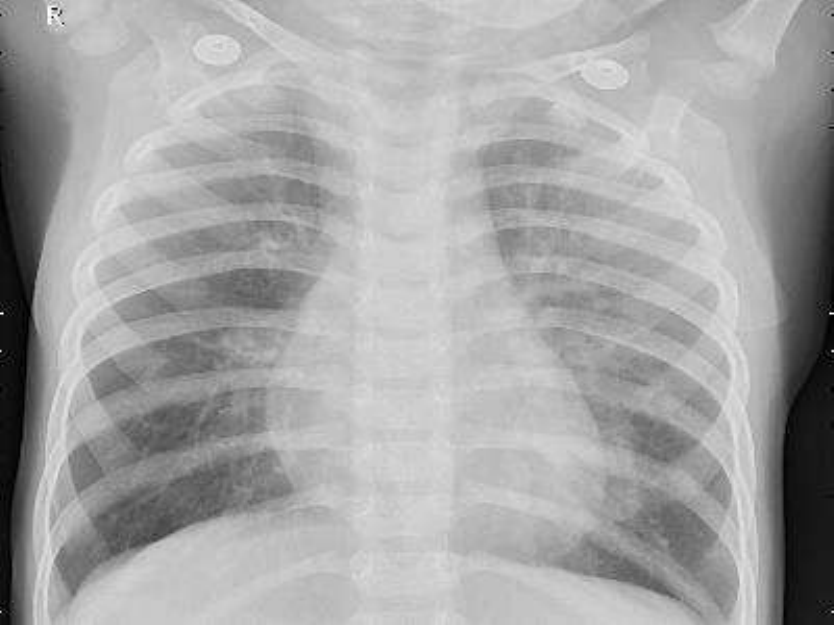}
	}
	\subfigure[COVID-19]{
		\label{COVID-19}
		\includegraphics[height=0.13\textwidth]{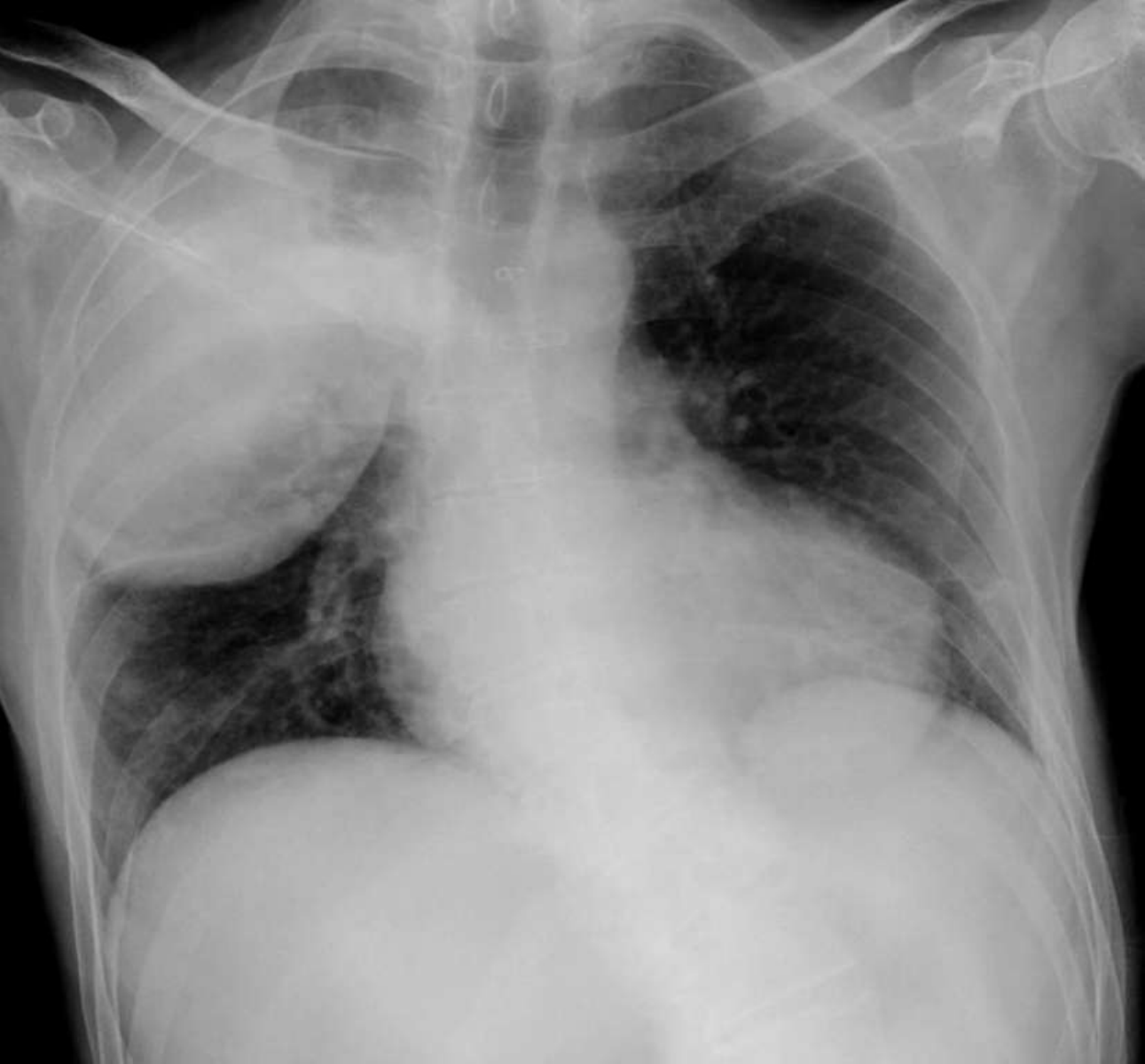}
	}
	\caption{Chest X-ray images.}
	\label{fig:covid19}
\end{figure}

Coronavirus disease 2019 (COVID-19) is a contagious disease caused by a virus~\cite{covid-19}.
According to Oxford English Dictionary~\cite{covid-19-d}, it is mainly characterized by fever and cough.
It is capable of progressing to pneumonia, respiratory and renal failure, blood coagulation abnormalities, and death, esp. in the elderly and people with underlying health conditions.
The first known case was identified in Wuhan, China, in December 2019, and then the disease quickly spread worldwide, resulting in the COVID-19 pandemic~\cite{covid-19}.
There exist several ways to diagnose COVID-19, including viral testing, imaging, coding, pathology, etc.
Chest CT scans may be one of the most useful ways to detect it~\cite{salehi2020coronavirus}.
\figurename~\ref{fig:c19-n}-\figurename~\ref{COVID-19} give an example of chest X-ray images.

\subsubsection{Experiment}

\begin{figure}[ht!]
	\centering
	\includegraphics[height=0.2\textwidth]{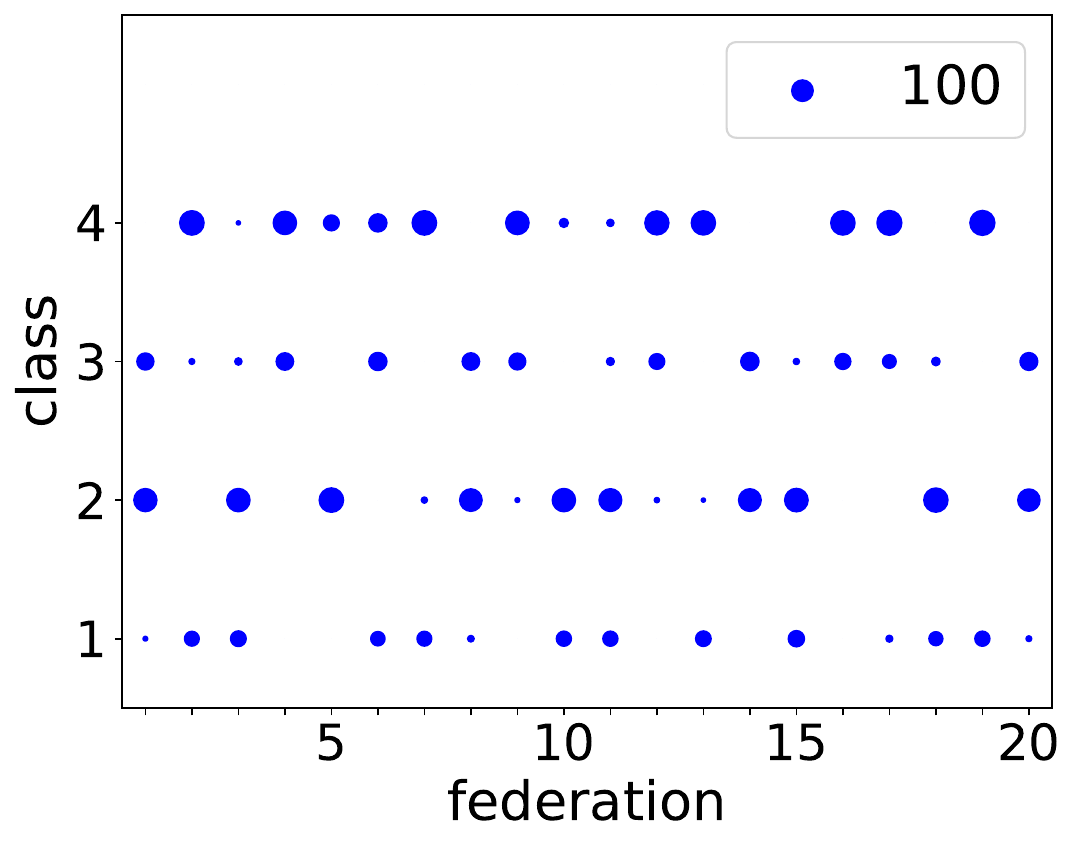}
	\caption{\lw{The number of samples per class allocated to each federation (indicated by dot size) on Covid 19.}}
	\label{fig:datasplit-covid}
\end{figure}

\paragraph{Dataset}
COVID-19 is a public COVID-19 posterior anterior chest radiography image dataset~\cite{sait2020curated}.
It is a combined curated Chest X-ray image dataset that collates 15 public datasets.
Four classes (1,281 COVID-19 X-Rays, 3,270 Normal X-Rays, 1,656 viral-pneumonia X-Rays, and 3,001 bacterial-pneumonia X-Rays) exist in the dataset and there are 9,208 images in total.
We utilize this public benchmark to evaluate our method with a real health-related application under label shifts.
Similar to Medical Image Classification, we utilize Dirichlet distribution to split data and \figurename~\ref{fig:datasplit-covid} visualizes how samples are distributed. 
In each federation, $40\%$, $30\%$, and $30\%$ of data are used for training, validation, and testing respectively.

\paragraph{Implementation Details}
For this real application, we utilized AlexNet, and other settings are similar to VLCS.

\paragraph{Results}
\input{tab-covid19}
The classification results for each federation on COVID-19 are shown in \tablename~\ref{tab:my-table-covid}.
We have the following observations from these results.
1) Our method achieves the best performance in this real health-related application. 
Compared to the second-best method, FedProx, \method has a significant improvement of $4.68\%$. 
Moreover, our method almost achieves the best performance on every federation.
2) In this real application under label shifts, FedBN does not achieve the second-best performance, since it is originally designed for feature shifts. 
Instead, FedProx achieves the second best while FedAvg even performs better than FedBN.

\subsection{Parkinson Disease}
\subsubsection{Background}


Parkinson's disease (Parkinson's, PD) is a long-term degenerative disorder of the central nervous system~\cite{pd-wiki}.
It leads a significant drop in the patient's quality of life and has drawn increasing attention for the substantial morbidity, increased mortality, and high economic burden~\cite{weintraub2008parkinson}.
Until now, there exist no disease-modifying drugs (drugs that target the causes or damage) are approved for Parkinson's.
Large amounts of costs are paid for PD, as a study based on 2017 data estimated that the US economic PD burden at \$51.9 billion, including direct medical costs of \$25.4 billion and \$26.5 billion in indirect and non-medical costs~\cite{yang2020current}.
According to \cite{jankovic2008parkinson}, patients with PD often suffer from four major motor symptoms, including tremor, rigidity, bradykinesia, and postural instability.

\subsubsection{PDAssist}

\begin{figure}[ht!]
	\centering
	\subfigure[Architecture]{
		\label{fig:pd2}
		\includegraphics[height=0.18\textwidth]{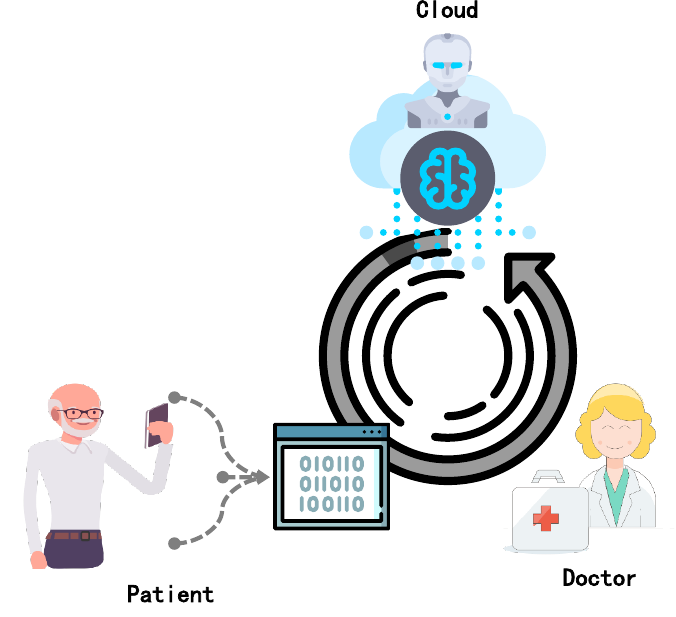}
	}
	\subfigure[Tremor testings]{
		\label{fig:pd3}
		\includegraphics[height=0.18\textwidth]{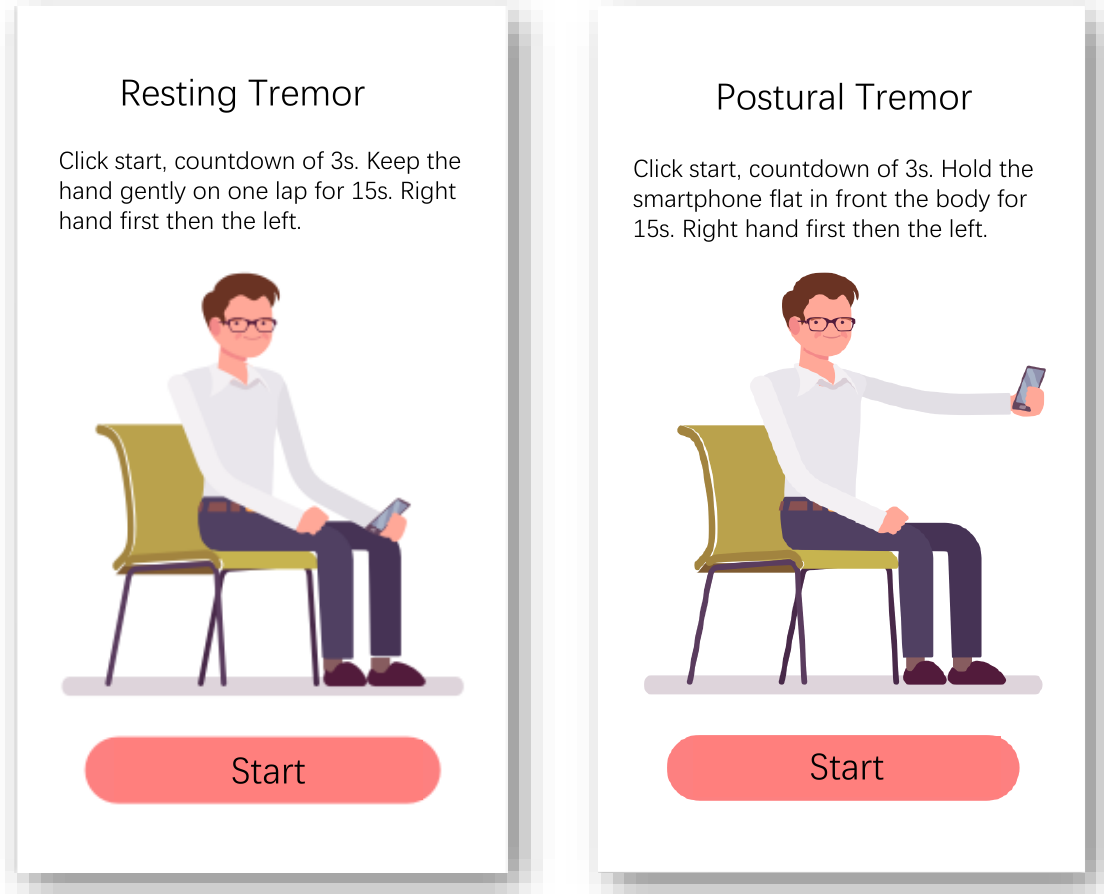}
	}
	\caption{PDAssist. \figurename~\ref{fig:pd2} is the architecture of PDAssist while \figurename~\ref{fig:pd3} describes tremor testings with PDAssist }
	\label{fig:pdas}
\end{figure} 


To reduce the pressure on doctors, our team constructed a mobile PD assessment tool, PDAssist~\cite{chen2017pdassist}.
PDAssist is based on the Unified Parkinson’s Disease Rating Scale (UPDRS)~\cite{movement2003unified} and it can assess PD symptoms automatically.
As shown in \figurename~\ref{fig:pd2}, the patient does some specially designed tasks.
At the same time, the mobile phone collects corresponding data and uploads them to the hospital where the patient belongs to.
The cloud in the corresponding hospital will offer an estimation score, from 0 to 4 (from normal to severe). 
Then the corresponding doctor can directly have an estimation of the patient's situation.
This tool can greatly reduce the pressure on doctors and brings patients a better medical experience.


\subsubsection{Experiment}
\paragraph{Dataset}
In this paper, we mainly focus on the tremor, a typical symptom of PD.
We collect data via PDAssist and the corresponding tasks are shown in \figurename~\ref{fig:pd3}.
We totally collect 11616 time-series samples from 167 patients.
Each sample contains 6 channels of two sensors, an accelerometer and a gyroscope, and the dimension is $6\times 32$.
Since 167 patients are distributed in 16 hospitals and each hospital can not directly exchange data with others, a federated learning method is required to construct personalized models for each hospital.
More notably, due to different hobbies, lifestyles, body shapes, devices, etc, there naturally exist feature shifts on data among different hospitals.
Simplified, in this task, we only predict whether the person is with PD.

\paragraph{Implementation Details}
For this real application, we utilize a CNN composed of two convolutional layers, two pooling layers, two batch normalization layers, and two fully connected layers. 
Please note the kernel sizes are different between this task and PAMAP2 and other settings are similar to VLCS.\looseness=-1

\paragraph{Results}
The classification results for each federation on Tremor of Parkinson's disease are shown in \tablename~\ref{tab:my-table-pd}.
We have the following observations from these results.
1) Our method still achieves the best performance in this real application. 
And it has improvements of $16.11\%$, $16.10\%$, and $6.48\%$ compared to FedAvg, FedProx, and FedBN respectively.
2) In this real application under feature shifts, FedBN achieves the second-best performance while FedAvg ad FedProx perform terribly.\looseness=-1

\input{tab-pd}

\subsection{Summary}
In these two real health-related applications, our method achieves the best performance compared to other state-of-the-art methods, which demonstrates the superiority of our method.
FedBN performs worst under label shifts different from Experiments (\sectionname~\ref{sec:exp}) while it still achieves the second best under feature shifts.

\section{Conclusion and Future Work}
\label{sec-conclus}
In this article, we proposed \method which uses cyclic knowledge distillation for meta federated learning.
\method organizes federations in another novel style that does not require a central server.
Comprehensive experiments have demonstrated the effectiveness of \method.
Moreover, \method has shown its superiority in real health-related applications.
In the future, we plan to combine \method with common methods such as FedAvg, to implement a complete federated learning system, including intra- and inter- federations. 
We also plan to apply \method for heterogeneity architectures and more realistic healthcare applications.

\bibliographystyle{IEEEtran}
\bibliography{IEEEabrv,refs}

\end{document}

%% file: tab-vlcs.tex
\begin{table}[htbp]
\centering
\caption{Accuracy on VLCS. Bold means the best results.}
\resizebox{0.45\textwidth}{!}{
\begin{tabular}{cccccc}
\toprule
method        & Caltech101     & LabelMe        & SUN09          & VOC2007        & AVG            \\ \midrule
FedAvg  & 82.69          & 54.43          & 51.52          & 44.89          & 58.38          \\
FedProx & 83.04          & 55.74          & 51.98          & \textbf{47.70} & 59.62          \\
FedBN   & 90.81          & 54.80          & 50.15          & 44.30          & 60.02          \\
\method & \textbf{94.35} & \textbf{59.32} & \textbf{57.62} & 45.78          & \textbf{64.27}
\\ \bottomrule
\end{tabular}}

\label{tab:my-table-vlcs}
\end{table}

%% file: tab-pacs.tex
\begin{table}[htbp]
\centering
\caption{Accuracy on PACS. Bold means the best results.}
\resizebox{0.45\textwidth}{!}{
\begin{tabular}{cccccc}
\toprule
Client                 & Art\_painting  & Cartoon        & Photo          & Sketch         & Avg            \\ \midrule
FedAvg                 & 29.34          & 59.40          & 53.59          & 60.25          & 50.65          \\
FedProx                & 33.01          & 63.46          & 50.90          & 61.02          & 52.10          \\
FedBN                  & 36.92          & 64.74          & 61.08          & 76.56          & 59.83          \\ 
\method & \textbf{43.03} & \textbf{69.44} & \textbf{63.77} & \textbf{80.00} & \textbf{64.06}\\ \bottomrule
\end{tabular}}

\label{tab:my-table-pacs}
\end{table}

%% file: tab-xp.tex
\begin{table}[htbp]
\centering
\caption{Accuracy on PAMAP2 for the Cross-Person setting. Bold means the best results.}
\resizebox{0.45\textwidth}{!}{
\begin{tabular}{cccccc}
\toprule
Method                 & 0              & 1              & 2              & 3              & AVG            \\ \midrule
FedAvg                 & 86.99          & 83.76          & 75.81          & 85.87          & 83.11          \\
FedProx                & 86.82          & 84.73          & 78.87          & \textbf{86.36} & 84.19          \\
FedBN                  & \textbf{89.19} & 85.53          & 80.00          & 85.39          & 85.03          \\
\method & 88.68          & \textbf{87.14} & \textbf{82.74} & 85.71          & \textbf{86.07}
\\ \bottomrule
\end{tabular}}

\label{tab:my-table-xp}
\end{table}

%% file: tab-pamap-tn.tex
\begin{table}[htbp]
\centering
\caption{Accuracy on PAMAP2. Bold means the best results.}
\resizebox{0.4\textwidth}{!}{
\begin{tabular}{ccccc}
\toprule
Client & FedAvg & FedProx & FedBN          & \method \\ \midrule
0      & 82.44  & 82.44   & 84.73          & \textbf{85.50}         \\
1      & 80.99  & 81.37   & 84.79          & \textbf{98.10}         \\
2      & 50.19  & 52.49   & 81.61          & \textbf{87.74}         \\
3      & 91.25  & 92.02   & \textbf{93.54} & 90.49                  \\
4      & 81.99  & 82.38   & 81.23          & \textbf{84.67}         \\
5      & 68.97  & 70.50   & \textbf{80.08} & 75.10                  \\
6      & 76.63  & 77.78   & 89.27          & \textbf{92.72}         \\
7      & 90.08  & 90.84   & 94.66          & \textbf{96.56}         \\
8      & 74.71  & 78.16   & 80.08          & \textbf{90.04}         \\
9      & 90.46  & 89.31   & 95.80          & \textbf{99.24}         \\
10     & 68.20  & 69.35   & 87.36          & \textbf{94.64}         \\
11     & 67.82  & 70.88   & \textbf{78.16} & 73.95                  \\
12     & 92.02  & 90.87   & 86.69          & \textbf{92.40}         \\
13     & 79.62  & 79.62   & \textbf{88.08} & 83.85                  \\
14     & 58.17  & 63.88   & 86.31          & \textbf{95.06}         \\
15     & 92.78  & 93.92   & \textbf{96.58} & 95.44                  \\
16     & 79.09  & 77.95   & \textbf{93.54} & \textbf{93.54}         \\
17     & 71.65  & 71.26   & 80.08          & \textbf{86.97}         \\
18     & 87.79  & 85.88   & \textbf{87.02} & 86.26                  \\
19     & 97.70  & 96.93   & \textbf{99.23} & \textbf{99.23}         \\
AVG    & 79.13  & 79.89   & 87.44          & \textbf{90.07}        \\ \bottomrule
\end{tabular}}

\label{tab:my-table-pamap}
\end{table}

%% file: tab-medmnist-tn.tex
\begin{table*}[htbp]
\caption{Accuracy on three benchmarks of MedMnist. Bold means the best result.}
\resizebox{\textwidth}{!}{
\begin{tabular}{c|cccc|cccc|cccc}
\toprule
Dataset & \multicolumn{4}{c|}{OrganA}                        & \multicolumn{4}{c|}{OrganC}                                        & \multicolumn{4}{c}{OrganS}                        \\
\midrule
Client  & FedAvg & FedProx & FedBN & \method & FedAvg & FedProx        & FedBN          & \method & FedAvg & FedProx & FedBN & \method \\
0       & 94.20  & 93.75   & 94.09 & \textbf{96.02}         & 86.40  & 87.25          & 82.44          & \textbf{90.65}         & 52.39  & 52.13   & 75.00 & \textbf{96.01}         \\
1       & 91.23  & 91.34   & 92.03 & \textbf{95.79}         & 90.00  & 90.29          & 94.86          & \textbf{96.29}         & 50.26  & 51.59   & 76.98 & \textbf{98.41}         \\
2       & 92.61  & 92.95   & 93.40 & \textbf{97.72}         & 88.03  & 85.76          & 92.02          & \textbf{93.16}         & 67.47  & 69.07   & 75.20 & \textbf{82.40}         \\
3       & 89.85  & 90.76   & 90.31 & \textbf{95.55}         & 88.39  & 87.82          & 91.50          & \textbf{94.05}         & 67.82  & 67.82   & 79.79 & \textbf{88.83}         \\
4       & 96.25  & 96.47   & 97.27 & \textbf{97.95}         & 84.05  & 85.47          & 86.32          & \textbf{88.89}         & 75.67  & 77.01   & 79.95 & \textbf{85.83}         \\
5       & 88.62  & 89.65   & 88.96 & \textbf{92.61}         & 90.31  & 89.74          & 92.88          & \textbf{90.60}         & 53.33  & 53.33   & 54.13 & \textbf{89.87}         \\
6       & 90.44  & 90.79   & 92.49 & \textbf{96.02}         & 78.35  & 79.20          & 78.92          & \textbf{82.62}         & 82.98  & 81.38   & 86.97 & \textbf{96.28}         \\
7       & 94.65  & 94.65   & 95.90 & \textbf{96.70}         & 95.16  & 96.58          & 97.44          & \textbf{97.72}         & 61.97  & 59.84   & 67.29 & \textbf{98.67}         \\
8       & 92.84  & 93.30   & 94.89 & \textbf{98.18}         & 83.52  & 84.09          & 86.36          & \textbf{90.34}         & 82.35  & 82.89   & 92.51 & \textbf{94.92}         \\
9       & 89.29  & 89.29   & 89.18 & \textbf{94.31}         & 91.45  & 90.88          & 90.03          & \textbf{95.73}         & 90.72  & 90.19   & 99.20 & \textbf{99.73}         \\
10      & 91.81  & 91.81   & 92.15 & \textbf{96.93}         & 85.76  & \textbf{86.89} & 85.19          & \textbf{86.89}         & 65.78  & 59.36   & 75.67 & \textbf{88.50}         \\
11      & 92.94  & 91.69   & 91.69 & \textbf{93.96}         & 90.03  & 91.74          & 90.60          & \textbf{94.02}         & 83.42  & 86.10   & 86.90 & \textbf{90.64}         \\
12      & 89.62  & 90.99   & 91.11 & \textbf{95.33}         & 86.04  & 87.18          & 83.76          & \textbf{89.17}         & 73.53  & 71.12   & 76.47 & \textbf{84.22}         \\
13      & 93.84  & 93.50   & 92.25 & \textbf{96.69}         & 90.63  & 91.76          & 92.90          & \textbf{95.74}         & 90.13  & 89.33   & 90.93 & \textbf{97.07}         \\
14      & 93.86  & 92.83   & 92.72 & \textbf{97.50}         & 86.36  & 87.50          & 85.80          & \textbf{91.76}         & 83.11  & 83.65   & 89.28 & \textbf{89.81}         \\
15      & 89.65  & 89.65   & 90.33 & \textbf{94.99}         & 91.43  & 92.29          & \textbf{94.29} & 93.71                  & 46.42  & 45.89   & 72.15 & \textbf{81.96}         \\
16      & 93.52  & 92.83   & 93.52 & \textbf{97.38}         & 86.89  & 87.75          & 88.03          & \textbf{93.16}         & 44.80  & 43.20   & 57.87 & \textbf{84.00}         \\
17      & 90.43  & 90.43   & 91.00 & \textbf{94.99}         & 92.33  & 90.91          & 92.90          & \textbf{95.74}         & 61.17  & 62.77   & 69.95 & \textbf{88.03}         \\
18      & 92.26  & 92.04   & 92.04 & \textbf{97.16}         & 85.76  & 86.04          & 82.34          & \textbf{87.18}         & 56.38  & 58.78   & 73.67 & \textbf{96.28}         \\
19      & 92.38  & 91.92   & 91.13 & \textbf{97.27}         & 91.76  & 92.33          & 92.90          & \textbf{94.03}         & 79.41  & 79.68   & 87.43 & \textbf{94.12}         \\
AVG     & 92.01  & 92.03   & 92.32 & \textbf{96.15}         & 88.13  & 88.57          & 89.07          & \textbf{92.07}         & 68.46  & 68.26   & 78.37 & \textbf{91.28}        
\\
\bottomrule
\end{tabular}}

\label{tab:my-table-medmnist}
\end{table*}

%% file: tab-covid19.tex
\begin{table}[htbp]
\centering
\caption{Accuracy on COVID-19. Bold means the best results.}
\resizebox{0.4\textwidth}{!}{
\begin{tabular}{ccccc}
\toprule
Client & FedAvg & FedProx & FedBN          & \method \\ \midrule
0&  74.45          & 76.64          & 70.07          & \textbf{78.10}  \\
1&  95.59          & 97.06          & 92.65          & \textbf{99.26}  \\
2&  87.50          & 85.29          & 91.18          & \textbf{91.91}  \\
3&  89.05          & 90.51          & 93.43          & \textbf{97.81}  \\
4&  88.32          & 89.05          & 95.62          & \textbf{97.08}  \\
5&  84.56          & 86.76          & 85.29          & \textbf{97.06}  \\
6&  97.81          & \textbf{98.54} & 93.43          & 97.81           \\
7&  73.53          & 75.74          & 69.12          & \textbf{80.88}  \\
8&  86.86          & 87.59          & \textbf{91.97} & \textbf{91.97}  \\
9&  87.50          & 89.71          & 94.12          & \textbf{100.00} \\
10&  83.94          & 83.94          & 87.59          & \textbf{88.32}  \\
11&  81.02          & 84.67          & 83.21          & \textbf{94.16}  \\
12&  \textbf{97.79} & 97.06          & 93.38          & \textbf{97.79}  \\
13&  \textbf{73.72} & \textbf{73.72} & 55.47          & 71.53           \\
14&  85.40          & 86.86          & 91.24          & \textbf{95.62}  \\
15&  84.56          & 88.97          & 91.18          & \textbf{97.79}  \\
16&  93.43          & 94.16          & 89.78          & \textbf{96.35}  \\
17&  86.13          & 87.59          & \textbf{89.78} & \textbf{89.78}  \\
18&  96.38          & 96.38          & 91.30          & \textbf{98.55}  \\
19&  75.91          & 75.91          & 64.96          & \textbf{78.10}  \\
AVG& 86.17          & 87.31          & 85.74          & \textbf{91.99}         \\ \bottomrule
\end{tabular}}

\label{tab:my-table-covid}
\end{table}

%% file: tab-pd.tex
\begin{table}[htbp]
\centering
\caption{Accuracy on Tremor of Parkinson's disease. Bold means the best results.}
\resizebox{0.4\textwidth}{!}{
\begin{tabular}{ccccc}
\toprule
Client & FedAvg & FedProx & FedBN          & \method \\ \midrule
0      & 60.13           & 61.44           & 86.27           & \textbf{91.50}         \\
1      & 85.47           & 85.47           & 79.49           & \textbf{97.44}         \\
2      & \textbf{100.00} & 99.68           & \textbf{100.00} & \textbf{100.00}        \\
3      & \textbf{100.00} & 96.95           & 99.24           & \textbf{100.00}        \\
4      & 84.17           & 81.30           & \textbf{89.93}  & 89.21                  \\
5      & 87.31           & 87.31           & \textbf{98.51}  & 96.27                  \\
6      & 56.55           & 56.55           & 61.38           & \textbf{73.10}         \\
7      & 59.40           & 59.40           & 58.65           & \textbf{75.94}         \\
8      & 44.14           & 44.14           & 64.83           & \textbf{66.21}         \\
9      & 40.30           & 45.52           & 69.40           & \textbf{79.85}         \\
10     & 50.37           & 50.37           & \textbf{75.91}  & \textbf{75.91}         \\
11     & 55.88           & 55.88           & 82.35           & \textbf{88.24}         \\
12     & 56.77           & 56.77           & 58.71           & \textbf{83.87}         \\
13     & 69.23           & 69.23           & 79.23           & \textbf{91.54}         \\
14     & \textbf{91.18}  & \textbf{91.18}  & \textbf{91.18}  & 89.71                  \\
15     & \textbf{100.00} & \textbf{100.00} & \textbf{100.00} & \textbf{100.00}        \\
AVG    & 71.31           & 71.32           & 80.94           & \textbf{87.42}       \\ \bottomrule
\end{tabular}}

\label{tab:my-table-pd}
\end{table}

%% file: metafed-tnnls.bbl
\begin{thebibliography}{10}
\providecommand{\url}[1]{#1}
\csname url@samestyle\endcsname
\providecommand{\newblock}{\relax}
\providecommand{\bibinfo}[2]{#2}
\providecommand{\BIBentrySTDinterwordspacing}{\spaceskip=0pt\relax}
\providecommand{\BIBentryALTinterwordstretchfactor}{4}
\providecommand{\BIBentryALTinterwordspacing}{\spaceskip=\fontdimen2\font plus
\BIBentryALTinterwordstretchfactor\fontdimen3\font minus
  \fontdimen4\font\relax}
\providecommand{\BIBforeignlanguage}[2]{{%
\expandafter\ifx\csname l@#1\endcsname\relax
\typeout{** WARNING: IEEEtran.bst: No hyphenation pattern has been}%
\typeout{** loaded for the language `#1'. Using the pattern for}%
\typeout{** the default language instead.}%
\else
\language=\csname l@#1\endcsname
\fi
#2}}
\providecommand{\BIBdecl}{\relax}
\BIBdecl

\bibitem{lu2022local}
W.~Lu, J.~Wang, and Y.~Chen, ``Local and global alignments for generalizable
  sensor-based human activity recognition,'' in \emph{IEEE International
  Conference on Acoustics, Speech and Signal Processing (ICASSP)}, 2022.

\bibitem{lu2022semantic}
W.~Lu, J.~Wang, Y.~Chen, S.~Pan, C.~Hu, and X.~Qin, ``Semantic-discriminative
  mixup for generalizable sensor-based cross-domain activity recognition,''
  \emph{IMWUT}, 2022.

\bibitem{he2021locality}
M.~He, J.~Zhang, S.~Shan, X.~Liu, Z.~Wu, and X.~Chen, ``Locality-aware
  channel-wise dropout for occluded face recognition,'' \emph{IEEE Transactions
  on Image Processing}, vol.~31, pp. 788--798, 2021.

\bibitem{lu2021cross}
W.~Lu, Y.~Chen, J.~Wang, and X.~Qin, ``Cross-domain activity recognition via
  substructural optimal transport,'' \emph{Neurocomputing}, vol. 454, pp.
  65--75, 2021.

\bibitem{li2021medical}
S.~Li, X.~Sui, X.~Luo, X.~Xu, Y.~Liu, and R.~S.~M. Goh, ``Medical image
  segmentation using squeeze-and-expansion transformers,'' in \emph{IJCAI},
  2021.

\bibitem{DBLP:conf/kdd/MaYLXS21}
\BIBentryALTinterwordspacing
F.~Ma, M.~Ye, J.~Luo, C.~Xiao, and J.~Sun, ``Advances in mining heterogeneous
  healthcare data,'' in \emph{{KDD} '21: The 27th {ACM} {SIGKDD} Conference on
  Knowledge Discovery and Data Mining, Virtual Event, Singapore, August 14-18,
  2021}, F.~Zhu, B.~C. Ooi, and C.~Miao, Eds.\hskip 1em plus 0.5em minus
  0.4em\relax {ACM}, 2021, pp. 4050--4051. [Online]. Available:
  \url{https://doi.org/10.1145/3447548.3470789}
\BIBentrySTDinterwordspacing

\bibitem{aguiar2022learning}
H.~Aguiar, M.~Santos, P.~Watkinson, and T.~Zhu, ``Learning of cluster-based
  feature importance for electronic health record time-series,'' in
  \emph{International Conference on Machine Learning}.\hskip 1em plus 0.5em
  minus 0.4em\relax PMLR, 2022, pp. 161--179.

\bibitem{liu2022contribution}
Z.~Liu, Y.~Chen, Y.~Zhao, H.~Yu, Y.~Liu, R.~Bao, J.~Jiang, Z.~Nie, Q.~Xu, and
  Q.~Yang, ``Contribution-aware federated learning for smart healthcare,'' in
  \emph{Proceedings of the 34th Annual Conference on Innovative Applications of
  Artificial Intelligence (IAAI-22)}, 2022.

\bibitem{inkster2018china}
N.~Inkster, \emph{China’s cyber power}.\hskip 1em plus 0.5em minus
  0.4em\relax Routledge, 2018.

\bibitem{voigt2017eu}
P.~Voigt and A.~Von~dem Bussche, ``The eu general data protection regulation
  (gdpr),'' \emph{A Practical Guide, 1st Ed., Cham: Springer International
  Publishing}, vol.~10, p. 3152676, 2017.

\bibitem{yang2019federated}
Q.~Yang, Y.~Liu, T.~Chen, and Y.~Tong, ``Federated machine learning: Concept
  and applications,'' \emph{ACM Transactions on Intelligent Systems and
  Technology (TIST)}, vol.~10, no.~2, pp. 1--19, 2019.

\bibitem{mcmahan2017communication}
B.~McMahan, E.~Moore, D.~Ramage, S.~Hampson, and B.~A. y~Arcas,
  ``Communication-efficient learning of deep networks from decentralized
  data,'' in \emph{Artificial Intelligence and Statistics}.\hskip 1em plus
  0.5em minus 0.4em\relax PMLR, 2017, pp. 1273--1282.

\bibitem{li2021fedbn}
X.~Li, M.~JIANG, X.~Zhang, M.~Kamp, and Q.~Dou, ``Fedbn: Federated learning on
  non-iid features via local batch normalization,'' in \emph{International
  Conference on Learning Representations}, 2021.

\bibitem{hinton2015distilling}
G.~Hinton, O.~Vinyals, J.~Dean \emph{et~al.}, ``Distilling the knowledge in a
  neural network,'' \emph{arXiv preprint arXiv:1503.02531}, vol.~2, no.~7,
  2015.

\bibitem{romero2014fitnets}
A.~Romero, N.~Ballas, S.~E. Kahou, A.~Chassang, C.~Gatta, and Y.~Bengio,
  ``Fitnets: Hints for thin deep nets,'' \emph{arXiv preprint arXiv:1412.6550},
  2014.

\bibitem{paluru2021anam}
N.~Paluru, A.~Dayal, H.~B. Jenssen, T.~Sakinis, L.~R. Cenkeramaddi, J.~Prakash,
  and P.~K. Yalavarthy, ``Anam-net: Anamorphic depth embedding-based
  lightweight cnn for segmentation of anomalies in covid-19 chest ct images,''
  \emph{IEEE Transactions on Neural Networks and Learning Systems}, vol.~32,
  no.~3, pp. 932--946, 2021.

\bibitem{ding2021deepkeygen}
Y.~Ding, F.~Tan, Z.~Qin, M.~Cao, K.-K.~R. Choo, and Z.~Qin, ``Deepkeygen: a
  deep learning-based stream cipher generator for medical image encryption and
  decryption,'' \emph{IEEE Transactions on Neural Networks and Learning
  Systems}, 2021.

\bibitem{yu2022healthnet}
F.~Yu, L.~Cui, H.~Chen, Y.~Cao, N.~Liu, W.~Huang, Y.~Xu, and H.~Lu,
  ``Healthnet: A health progression network via heterogeneous medical
  information fusion,'' \emph{IEEE Transactions on Neural Networks and Learning
  Systems}, 2022.

\bibitem{wang2019deep}
J.~Wang, Y.~Chen, S.~Hao, X.~Peng, and L.~Hu, ``Deep learning for sensor-based
  activity recognition: A survey,'' \emph{Pattern Recognition Letters}, vol.
  119, pp. 3--11, 2019.

\bibitem{9129779}
K.~Muhammad, S.~Khan, J.~D. Ser, and V.~H. C.~d. Albuquerque, ``Deep learning
  for multigrade brain tumor classification in smart healthcare systems: A
  prospective survey,'' \emph{IEEE Transactions on Neural Networks and Learning
  Systems}, vol.~32, no.~2, pp. 507--522, 2021.

\bibitem{tan2022towards}
A.~Z. Tan, H.~Yu, L.~Cui, and Q.~Yang, ``Towards personalized federated
  learning,'' \emph{IEEE Transactions on Neural Networks and Learning Systems},
  2022.

\bibitem{vogenberg2010personalized}
F.~R. Vogenberg, C.~I. Barash, and M.~Pursel, ``Personalized medicine: part 1:
  evolution and development into theranostics,'' \emph{Pharmacy and
  Therapeutics}, vol.~35, no.~10, p. 560, 2010.

\bibitem{li2018federated}
T.~Li, A.~K. Sahu, M.~Zaheer, M.~Sanjabi, A.~Talwalkar, and V.~Smith,
  ``Federated optimization in heterogeneous networks,'' \emph{Proceedings of
  Machine Learning and Systems}, vol.~2, pp. 429--450, 2020.

\bibitem{yu2020salvaging}
T.~Yu, E.~Bagdasaryan, and V.~Shmatikov, ``Salvaging federated learning by
  local adaptation,'' \emph{arXiv preprint arXiv:2002.04758}, 2020.

\bibitem{chenbridging}
H.-Y. Chen and W.-L. Chao, ``On bridging generic and personalized federated
  learning for image classification,'' in \emph{International Conference on
  Learning Representations}, 2022.

\bibitem{chen2020fedhealth}
Y.~Chen, X.~Qin, J.~Wang, C.~Yu, and W.~Gao, ``Fedhealth: A federated transfer
  learning framework for wearable healthcare,'' \emph{IEEE Intelligent
  Systems}, vol.~35, no.~4, pp. 83--93, 2020.

\bibitem{lu2022personalized}
W.~Lu, J.~Wang, Y.~Chen, X.~Qin, R.~Xu, D.~Dimitriadis, and T.~Qin,
  ``Personalized federated learning with adaptive batchnorm for healthcare,''
  \emph{IEEE Transactions on Big Data}, 2022.

\bibitem{kopparapu2020fedfmc}
K.~Kopparapu and E.~Lin, ``Fedfmc: Sequential efficient federated learning on
  non-iid data,'' \emph{arXiv preprint arXiv:2006.10937}, 2020.

\bibitem{zaccone2022speeding}
R.~Zaccone, A.~Rizzardi, D.~Caldarola, M.~Ciccone, and B.~Caputo, ``Speeding up
  heterogeneous federated learning with sequentially trained superclients,''
  \emph{arXiv}, 2022.

\bibitem{zeng2022heterogeneous}
S.~Zeng, Z.~Li, H.~Yu, Y.~He, Z.~Xu, D.~Niyato, and H.~Yu, ``Heterogeneous
  federated learning via grouped sequential-to-parallel training,'' in
  \emph{Database Systems for Advanced Applications - 27th International
  Conference, {DASFAA} 2022, Virtual Event, April 11-14, 2022, Proceedings,
  Part{II}}, ser. Lecture Notes in Computer Science, vol. 13246.\hskip 1em plus
  0.5em minus 0.4em\relax Springer, 2022, pp. 455--471.

\bibitem{roy2019braintorrent}
A.~G. Roy, S.~Siddiqui, S.~P{\"o}lsterl, N.~Navab, and C.~Wachinger,
  ``Braintorrent: A peer-to-peer environment for decentralized federated
  learning,'' \emph{arXiv}, 2019.

\bibitem{rieke2020future}
N.~Rieke, J.~Hancox, W.~Li, F.~Milletari, H.~R. Roth, S.~Albarqouni, S.~Bakas,
  M.~N. Galtier, B.~A. Landman, K.~Maier-Hein \emph{et~al.}, ``The future of
  digital health with federated learning,'' \emph{NPJ digital medicine},
  vol.~3, no.~1, pp. 1--7, 2020.

\bibitem{li2021fedh2l}
Y.~Li, W.~Zhou, H.~Wang, H.~Mi, and T.~M. Hospedales, ``Fedh2l: Federated
  learning with model and statistical heterogeneity,'' \emph{arXiv}, 2021.

\bibitem{warnat2021swarm}
S.~Warnat-Herresthal, H.~Schultze, K.~L. Shastry, S.~Manamohan, S.~Mukherjee,
  V.~Garg, R.~Sarveswara, K.~H{\"a}ndler, P.~Pickkers, N.~A. Aziz
  \emph{et~al.}, ``Swarm learning for decentralized and confidential clinical
  machine learning,'' \emph{Nature}, vol. 594, no. 7862, pp. 265--270, 2021.

\bibitem{zhang2021survey}
J.~Zhang, X.~Yang, H.~Meng, Z.~Lin, Y.~Xu, and L.~Cui, ``A survey on knowledge
  enhanced ehr data mining,'' in \emph{5th International Conference on Crowd
  Science and Engineering}, 2021, pp. 124--131.

\bibitem{meng2021knowledge}
H.~Meng, Z.~Lin, F.~Yang, Y.~Xu, and L.~Cui, ``Knowledge distillation in
  medical data mining: a survey,'' in \emph{5th International Conference on
  Crowd Science and Engineering}, 2021, pp. 175--182.

\bibitem{usmanova2021distillation}
A.~Usmanova, F.~Portet, P.~Lalanda, and G.~Vega, ``A distillation-based
  approach integrating continual learning and federated learning for pervasive
  services,'' in \emph{3rd Workshop on Continual and Multimodal Learning for
  Internet of Things--Co-located with IJCAI 2021}, 2021.

\bibitem{afonin2021towards}
A.~Afonin and S.~P. Karimireddy, ``Towards model agnostic federated learning
  using knowledge distillation,'' in \emph{International Conference on Learning
  Representations}, 2022.

\bibitem{fang2013unbiased}
C.~Fang, Y.~Xu, and D.~N. Rockmore, ``Unbiased metric learning: On the
  utilization of multiple datasets and web images for softening bias,'' in
  \emph{Proceedings of the IEEE International Conference on Computer Vision},
  2013, pp. 1657--1664.

\bibitem{krizhevsky2012imagenet}
A.~Krizhevsky, I.~Sutskever, and G.~E. Hinton, ``Imagenet classification with
  deep convolutional neural networks,'' in \emph{NeurIPS}, vol.~25, 2012, pp.
  1097--1105.

\bibitem{li2017deeper}
D.~Li, Y.~Yang, Y.-Z. Song, and T.~M. Hospedales, ``Deeper, broader and artier
  domain generalization,'' in \emph{Proceedings of the IEEE international
  conference on computer vision}, 2017, pp. 5542--5550.

\bibitem{reiss2012introducing}
A.~Reiss and D.~Stricker, ``Introducing a new benchmarked dataset for activity
  monitoring,'' in \emph{2012 16th International Symposium on Wearable
  Computers}.\hskip 1em plus 0.5em minus 0.4em\relax IEEE, 2012, pp. 108--109.

\bibitem{lu2022domain}
W.~Lu, J.~Wang, H.~Li, Y.~Chen, and X.~Xie, ``Domain-invariant feature
  exploration for domain generalization,'' \emph{arXiv preprint
  arXiv:2207.12020}, 2022.

\bibitem{yurochkin2019bayesian}
M.~Yurochkin, M.~Agarwal, S.~Ghosh, K.~Greenewald, N.~Hoang, and Y.~Khazaeni,
  ``Bayesian nonparametric federated learning of neural networks,'' in
  \emph{ICML}, 2019, pp. 7252--7261.

\bibitem{bilic2019liver}
P.~Bilic, P.~Christ, H.~B. Li, E.~Vorontsov, A.~Ben-Cohen, G.~Kaissis,
  A.~Szeskin, C.~Jacobs, G.~E.~H. Mamani, G.~Chartrand \emph{et~al.}, ``The
  liver tumor segmentation benchmark (lits),'' \emph{Medical Image Analysis},
  p. 102680, 2022.

\bibitem{xu2019efficient}
X.~Xu, F.~Zhou, B.~Liu, D.~Fu, and X.~Bai, ``Efficient multiple organ
  localization in ct image using 3d region proposal network,'' \emph{IEEE
  transactions on medical imaging}, vol.~38, no.~8, pp. 1885--1898, 2019.

\bibitem{medmnistv1}
J.~Yang, R.~Shi, and B.~Ni, ``Medmnist classification decathlon: A lightweight
  automl benchmark for medical image analysis,'' in \emph{ISBI}, 2021, pp.
  191--195.

\bibitem{medmnistv2}
J.~Yang, R.~Shi, D.~Wei, Z.~Liu, L.~Zhao, B.~Ke, H.~Pfister, and B.~Ni,
  ``Medmnist v2: A large-scale lightweight benchmark for 2d and 3d biomedical
  image classification,'' \emph{arXiv preprint arXiv:2008.\#TODO}, 2021.

\bibitem{lecun1998gradient}
Y.~LeCun, L.~Bottou, Y.~Bengio, and P.~Haffner, ``Gradient-based learning
  applied to document recognition,'' \emph{Proceedings of the IEEE}, vol.~86,
  no.~11, pp. 2278--2324, 1998.

\bibitem{covid-19}
\BIBentryALTinterwordspacing
WIKIPEDIA. {COVID-19}. (2021, Feburary). [Online]. Available:
  \url{https://en.wikipedia.org/wiki/COVID-19#cite_ref-WSJ-20210226_7-0}
\BIBentrySTDinterwordspacing

\bibitem{covid-19-d}
\BIBentryALTinterwordspacing
O.~U. Press. {Oxford English Dictionary}. (2020, April). [Online]. Available:
  \url{https://www.oed.com/view/Entry/88575495}
\BIBentrySTDinterwordspacing

\bibitem{salehi2020coronavirus}
S.~Salehi, A.~Abedi, S.~Balakrishnan, A.~Gholamrezanezhad \emph{et~al.},
  ``Coronavirus disease 2019 (covid-19): a systematic review of imaging
  findings in 919 patients,'' \emph{Ajr Am J Roentgenol}, vol. 215, no.~1, pp.
  87--93, 2020.

\bibitem{sait2020curated}
U.~Sait, K.~Lal, S.~Prajapati, R.~Bhaumik, T.~Kumar, S.~Sanjana, and K.~Bhalla,
  ``Curated dataset for covid-19 posterior-anterior chest radiography images
  (x-rays),'' \emph{Mendeley Data}, vol.~1, 2020.

\bibitem{pd-wiki}
\BIBentryALTinterwordspacing
WIKIPEDIA. {Parkinson's disease}. (2021, April). [Online]. Available:
  \url{https://en.wikipedia.org/wiki/Parkinson\%27s_disease}
\BIBentrySTDinterwordspacing

\bibitem{weintraub2008parkinson}
D.~Weintraub, C.~L. Comella, and S.~Horn, ``Parkinson's disease--part 1:
  Pathophysiology, symptoms, burden, diagnosis, and assessment,'' \emph{Am J
  Manag Care}, vol.~14, no. 2 Suppl, pp. S40--S48, 2008.

\bibitem{yang2020current}
W.~Yang, J.~L. Hamilton, C.~Kopil, J.~C. Beck, C.~M. Tanner, R.~L. Albin,
  E.~Ray~Dorsey, N.~Dahodwala, I.~Cintina, P.~Hogan \emph{et~al.}, ``Current
  and projected future economic burden of parkinson’s disease in the us,''
  \emph{npj Parkinson's Disease}, vol.~6, no.~1, pp. 1--9, 2020.

\bibitem{jankovic2008parkinson}
J.~Jankovic, ``Parkinson’s disease: clinical features and diagnosis,''
  \emph{Journal of neurology, neurosurgery \& psychiatry}, vol.~79, no.~4, pp.
  368--376, 2008.

\bibitem{chen2017pdassist}
Y.~Chen, X.~Yang, B.~Chen, C.~Miao, and H.~Yu, ``Pdassist: Objective and
  quantified symptom assessment of parkinson's disease via smartphone,'' in
  \emph{2017 IEEE International Conference on Bioinformatics and Biomedicine
  (BIBM)}.\hskip 1em plus 0.5em minus 0.4em\relax IEEE, 2017, pp. 939--945.

\bibitem{movement2003unified}
M.~D. S. T.~F. on~Rating Scales~for Parkinson's~Disease, ``The unified
  parkinson's disease rating scale (updrs): status and recommendations,''
  \emph{Movement Disorders}, vol.~18, no.~7, pp. 738--750, 2003.

\end{thebibliography}
